\PassOptionsToPackage{numbers,square,sort&compress}{natbib}
\PassOptionsToPackage{table,dvipsnames}{xcolor}
\documentclass[10pt]{style/company_light}
\usepackage[utf8]{inputenc}
\usepackage{amsmath,amssymb,amsfonts}
\usepackage{tabularx,longtable,array,enumitem,adjustbox,needspace,float}
\usepackage{xurl}
\usepackage{upquote}
\definecolor{questionframe}{HTML}{4C42F5}
\definecolor{questionbackground}{HTML}{F3F2FF}
\newtcolorbox{questionbox}{colback=questionbackground,colframe=questionframe,
  coltext=black,boxrule=0.8pt,arc=3pt,left=9pt,right=9pt,top=5pt,bottom=5pt,
  before skip=7pt,after skip=8pt}

\newcolumntype{Y}{>{\raggedright\arraybackslash}X}

\newcommand{\method}{InsightPro}

\newtcolorbox{promptbox}[1]{breakable,colback=brandbg,colframe=brandprimary,
  title={#1},fonttitle=\bfseries\sffamily,coltitle=brandfg,colbacktitle=brandbg,
  boxrule=0.4pt,arc=2pt,left=7pt,right=7pt,top=5pt,bottom=5pt,
  before skip=8pt,after skip=8pt}
\newtcolorbox{draftbox}{colback=gray!5,colframe=gray!35,boxrule=.4pt,
  arc=2pt,left=7pt,right=7pt,top=5pt,bottom=5pt,fontupper=\small}
\hypersetup{pdftitle={InsightPro: Condensed Domain Knowledge as a First Step toward Autonomous Research},
  pdfauthor={},pdfsubject={InsightPro database and ideation prompts}}
\setlist[itemize]{leftmargin=1.3em,itemsep=2pt,topsep=4pt}
\setlist[enumerate]{leftmargin=1.6em,itemsep=2pt,topsep=4pt}

\makeatletter\@clubpenalty=10000\makeatother
\newcommand{\secgap}{\vspace{-2mm}}        
\newcommand{\figcapgap}{\par\vspace{-2mm}} 
\newcommand{\tabcaptop}{\setlength{\abovecaptionskip}{1mm}} 
\newcommand{\floatgap}{\vspace{-1mm}}      
\SetEnumitemKey{mainlist}{topsep=2pt,itemsep=0pt} 
\title{InsightPro: Condensed Domain Knowledge\\as a First Step toward Autonomous Research}
\author[*,\dagger]{Hanzuo Liu}
\author[*]{Kaicheng Yang}
\author[*]{Junfeng Tian}
\author[*]{Defeng Zhou}
\author[*]{Zhihan Shan}
\author{Xianglong Yan}
\author{Yibai Chen}
\author{Xin Yan}
\author{Wenyi Li}
\author{Chunyu Liu}
\author{Yizhe Chi}
\author{Yong Yan}
\author{Youjie Zheng}
\author{Zheng Jiang}
\author[\ddagger]{Calvin Xiao}
\author[\ddagger]{Kaisen Yang}
\author[\ddagger]{Huan-ang Gao}

\affiliation{Navers Lab, Einsia.AI,\quad\quad Tsinghua University}
\contribution[*]{Equal Contribution}
\contribution[\dagger]{Project Lead}
\contribution[\ddagger]{Corresponding Authors}

\renewcommand{\metadatalist}{%
{\small
\begin{tabularx}{\linewidth}{@{}>{\hsize=.88\hsize}X@{\hspace{8pt}}>{\hsize=1.12\hsize}X@{}}
\textsf{\textbf{Homepage:}} \url{https://lab.einsia.ai/insightpro} & \textsf{\textbf{Demo:}} \url{https://insightpro-demo.einsia.ai} \\
\textsf{\textbf{Code:}} \url{https://github.com/Einsia/InsightPro} & \textsf{\textbf{Dataset:}} \url{https://huggingface.co/datasets/Einsia/InsightPro} \\
\textsf{\textbf{Correspondence:}} \email{nana@einsia.ai} & \textsf{\textbf{Date:}} December 1, 2026
\end{tabularx}\par}
}
\abstract{%
Research ideation is a pivotal first step toward autonomous scientific discovery.
Human researchers often identify important problems in accumulated literature knowledge and connect insights within or across fields to propose new explanations or mechanisms for solving problems.
Inspired by this process, we combine reusable external knowledge organization with reasoning in an agent's limited working context.
We introduce \textbf{InsightPro}, a literature database that synthesizes AI research into compact insights and problems linked to supporting papers, enabling agents to access a field's accumulated research knowledge before ideation.
We use this resource to guide agents in selecting problems, connecting insights, developing mechanisms, and checking contributions against prior work.
Across 38 research domains and five generation models, single-agent and multi-agent prompting workflows based on this database lead adapted research ideation baselines in contribution, importance, novelty, and literature novelty under two independent LLM judges, and are in the first tier in soundness and proposal quality.
These results support a human-inspired approach to automated research ideation: organizing accumulated knowledge into an inspectable external structure and using it to guide problem selection and mechanism design.
}

\begin{document}
\typeout{M2P-LAYOUT textwidth=\the\textwidth\space columnwidth=\the\columnwidth\space textheight=\the\textheight}
\maketitle
{\centering\noindent\makebox[\linewidth][c]{\includegraphics[trim=6.95bp 6.95bp 6.95bp 6.95bp,clip,width=\dimexpr\linewidth+1.6cm-1mm\relax]{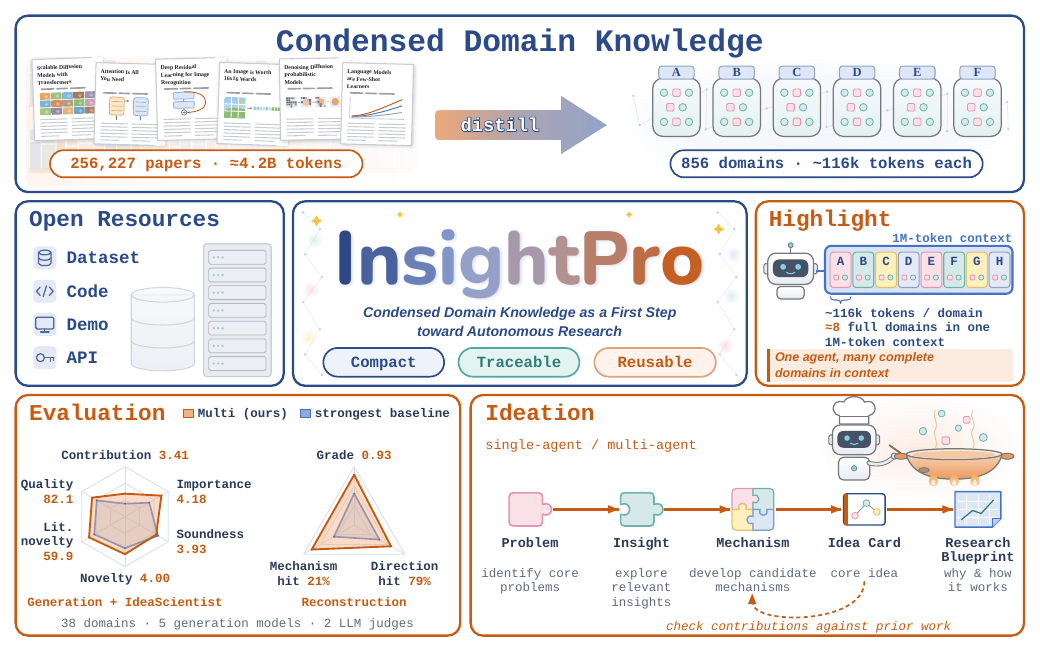}}\par}
\vspace{4pt}
\secgap\section{Introduction}
\label{sec:intro}
\vspace{-1.5mm}

Research ideation is an important starting point for research: it determines which problems to investigate, which mechanisms to propose, and which claims subsequent experiments must test, setting the direction of the entire research process~\citep{si2024ideation,liu2026ideascientist}.
This stage still depends heavily on human researchers' accumulated domain knowledge and judgment about important problems and potential breakthroughs.
As language models become more capable and agent context windows expand, agents are taking on more tasks in experimental execution and result analysis~\citep{aiscientist2024,dolphin2025}.
Enabling them to propose worthwhile research problems and well-grounded new mechanisms is therefore a key challenge for autonomous scientific discovery systems, and an important step for automated research toward forming research directions autonomously.
High-quality ideation requires retrieving, understanding, and synthesizing substantial literature to identify important problems and establish new connections among existing findings.
Its quality thus depends on both the model's reasoning ability and the density and accuracy of the knowledge available within its limited context.

\looseness=-1 Existing research ideation systems support agent ideation by improving literature access and organization.
For example, IdeaScientist retrieves structured research records to identify research gaps and transferable mechanisms~\citep{liu2026ideascientist}; Chain-of-Ideas (CoI) organizes related papers into research-development chains to trace the evolution of prior work~\citep{li2024coi}.
However, retrieving relevant papers does not by itself provide a knowledge representation suited to research ideation.
Repeated background and implementation details in complete papers consume limited context, whereas mechanisms, findings, and limitations shared across studies can often be expressed more concisely.
Content unrelated to the core problem can interfere with reasoning~\citep{shi2023distracted}; overstated findings and insufficiently specified applicability conditions can also mislead an agent's judgment.
If implementation difficulty becomes the main selection criterion too early in ideation, the agent may favor readily executable approaches, restricting its exploration of important problems and new mechanisms.
Moreover, retrieval records and a final proposal do not automatically expose the reasoning chain from literature evidence through borrowed insights to mechanism design, making it difficult for researchers to trace key assumptions and provide targeted critique.

\Needspace{6\baselineskip}
Human research practice offers inspiration for these challenges.
Researchers can draw on accumulated knowledge within limited working memory in part because they distill extensive reading and research experience into concepts and mechanisms that can be recalled and transferred~\citep{cowan2001capacity,ericsson1995memory}.
In this sense, \textbf{knowledge compression and organization are an important foundation of human research intelligence.}
This compression preserves relationships among findings, the conditions under which mechanisms hold, and unresolved obstacles, allowing a concise concept to evoke rich related knowledge.
Drawing on human research ideation therefore calls for compact, reliable, and traceable knowledge representations that give agents more useful domain knowledge for problem selection and mechanism design within limited context.

\begin{figure}[tp]
    \vspace{-2mm}\centering
    \includegraphics[width=1\linewidth]{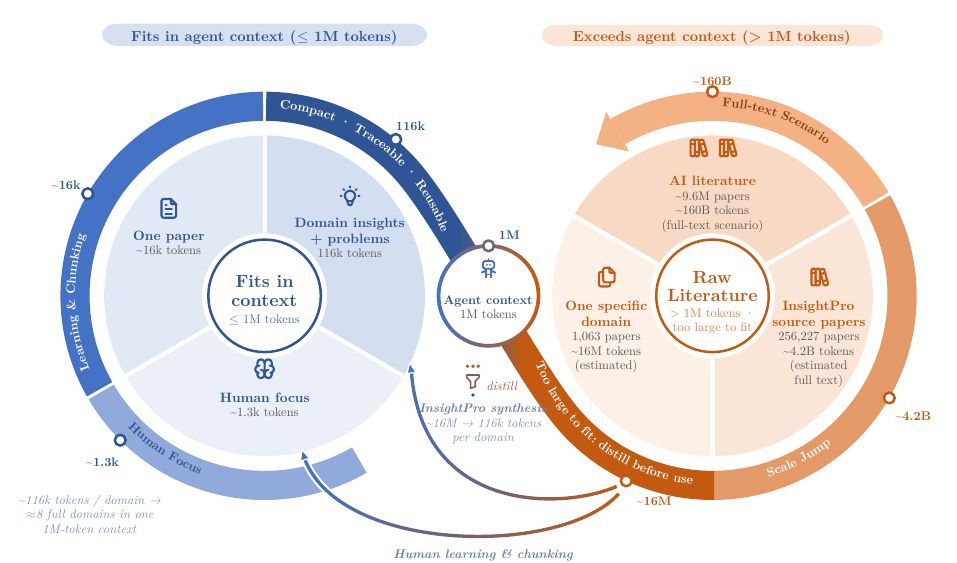}
    \figcapgap
    \caption{\textbf{Organizing literature for a limited working context.} Extrapolating the sampled PDF lengths gives approximately 4.2B full-text tokens for InsightPro's 256,227 source papers. The arrow labelled \emph{InsightPro synthesis} illustrates synthesis within one domain, whose exported insights and problems occupy approximately 116k tokens. Full-text totals are extrapolated from 180 sampled papers (Appendix~\ref{app:context-budget}).}
    \label{fig:zh-1}
    \label{fig:context-axis}
\floatgap
\end{figure}

To this end, we build \textbf{InsightPro}, a literature knowledge base for research ideation.
We collect 256,227 AI papers from the past decade, including major conferences such as NeurIPS, ICML, ICLR, CVPR, and ACL, organize them into 856 research domains, and synthesize domain knowledge into two core types of entry: \emph{insights} capture reusable findings, mechanisms, and their applicability conditions; \emph{problems} describe important obstacles, unresolved questions, and the limitations of existing methods.
Each entry retains links to its source papers, allowing an agent to compare compact research knowledge first and return to the original papers when it needs to verify details.
Through cross-paper synthesis, InsightPro reduces the context occupied by repeated content while preserving the sources and applicability boundaries that support judgment.

Building on InsightPro, we design ideation prompts that guide agents through a problem--insight--mechanism process.
Agents retrieve domain knowledge through an interface; we encourage agents to develop a thorough understanding of a domain's accumulated insights and problems before forming candidate ideas, then establish within-domain or cross-domain connections around important problems, propose mechanisms, and check their remaining contributions against prior work.
The prompts also require agents to record the evidence they use, key assumptions, and reasoning, and to deliver an Idea Card stating the core idea; a selected card can then be elaborated into a Research Blueprint describing its motivation and design, supporting further researcher inspection and refinement.

Our main contributions are:
\begin{enumerate}[leftmargin=*,itemsep=4pt,mainlist]
    \item \looseness=-1 \textbf{InsightPro: a compact literature database for research ideation.}
    We organize knowledge spread across many papers into a multilevel domain--insight/problem--source-paper structure, synthesizing shared findings and obstacles while keeping sources and applicability conditions.
    We release the database, complete-domain exports, and a retrieval interface to provide research knowledge that agents can compare, trace, and verify.
    \item \looseness=-1 \textbf{Applications and evaluation of InsightPro for research ideation.}
    We design generation prompts based on this database and execute them with single and multiple agents, guiding agents to connect problems with insights, develop mechanisms, and check prior work while retaining traceable research records.
    We evaluate generated proposals across 38 domains and combine historical contribution reconstruction with paired Card--Blueprint analyses to examine how this knowledge and its use support ideation and proposal development.
\end{enumerate}

\secgap\section{Methodology}
\label{sec:zh-method}
\label{sec:method}
\looseness=-1 Our goal is to improve the ideation stage of automated research by drawing on how human researchers organize prior knowledge and use it to develop new ideas. We construct the InsightPro database to provide reusable literature knowledge and design prompts that guide agents in using it to formulate and examine research proposals.

\Needspace{10\baselineskip}
\secgap\subsection{Research Motivation}
\label{sec:zh-human-ideation}
\label{sec:human-ideation}
\label{sec:ideation-challenges}

\begin{questionbox}
\textbf{Question 1:} How do humans develop research ideas?
\end{questionbox}

Through sustained reading, comparison, and research practice, researchers distill individual studies into concepts and mechanisms that they can recall, drawing on this accumulated knowledge during ideation~\citep{cowan2001capacity,ericsson1995memory}. Understanding established findings, their conditions of validity, and unresolved obstacles helps researchers judge which problems are worth solving and identify insights that could address them. New directions can emerge from recombining findings within a field or from mechanisms in other fields with similar structures. Researchers then examine the conditions under which a connection holds, compare it with close prior work, and revise an idea when they find duplication or gaps in its logic, gradually developing testable research claims. We derive design implications for agents from these recurring activities.

\Needspace{5\baselineskip}
\noindent\textbf{Observation 1: Compact domain knowledge underpins research ideation within a limited context.}

\looseness=-1 Agents likewise need to understand a field's established findings and obstacles, yet cannot readily retain the full texts of an entire field in their working context. Mechanisms and limitations that recur across papers can be synthesized, while the objects of study, conditions of validity, and exceptions must be retained to support subsequent judgments. A knowledge representation for ideation should therefore emphasize reusable insights and unresolved problems, linking them to source papers so that agents can first compare compact knowledge and then verify specific evidence as needed. InsightPro organizes these entries and their sources by domain (Section~\ref{sec:zh-insightpro}).

\Needspace{5\baselineskip}
\begin{questionbox}
\textbf{Question 2:} How can agents develop valuable new mechanisms from existing knowledge?
\end{questionbox}

\looseness=-1 After obtaining relevant material, an agent must still determine which problems matter, which insights could address the target obstacle, and what contribution remains after drawing on existing mechanisms. Retrieval from a short task description may miss important problems or relevant findings expressed in other terms; even topically relevant results may lack a useful mechanistic connection. Ideation also requires examining relationships among the objects of study, constraints, and operations, and judging whether they can support a new design.

\Needspace{5\baselineskip}
\noindent\textbf{Observation 2: Connecting insights around key problems provides a basis for mechanism design.}

\looseness=-1 Condensed domain insights and problems provide a reference for problem selection and supply findings relevant to specific obstacles. For a selected problem, the agent must explain why an insight is useful, which conditions of the original mechanism remain valid, and what changes are needed for transfer or combination. Problem selection and mechanism exploration can inform one another and be revised repeatedly as evidence changes. Accordingly, the InsightPro-based generation prompt guides agents to explore connections within and across domains, explain their reasoning from insight to design, and examine the remaining contribution through comparisons with close prior work (Section~\ref{sec:zh-ideax}).

\begin{figure}[tp]
\centering
\includegraphics[width=\linewidth]{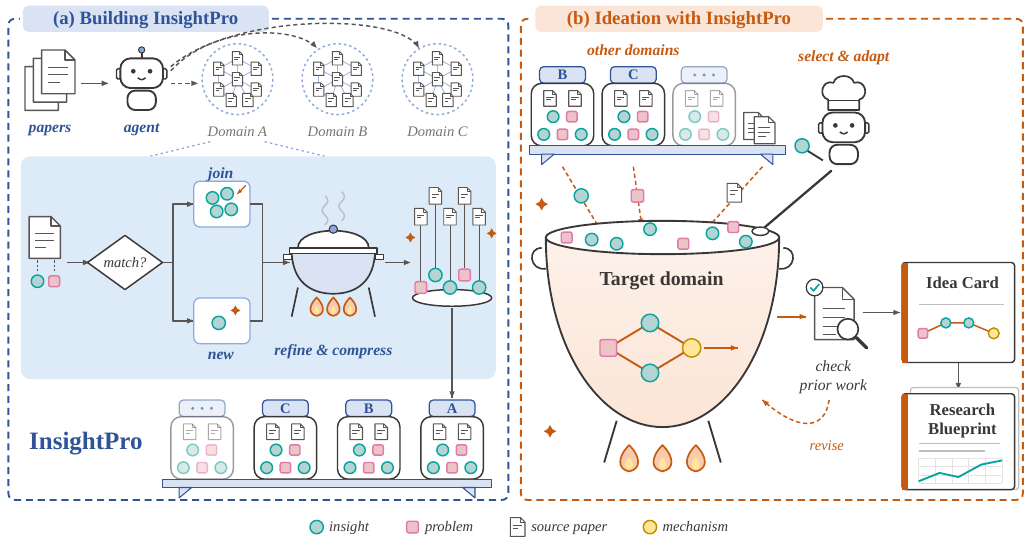}
\figcapgap
\caption{\textbf{From condensed knowledge to a research proposal.} (a)~An agent organizes papers into research domains; within a domain, each paper's findings and obstacles either join an existing entry about the same research object or start a new one, and are then refined and compressed into insights and problems linked to source papers. (b)~Agents retrieve relevant knowledge through an interface. A core problem guides exploration, while within-domain and cross-domain insights inform a proposed mechanism. Comparison with close prior work can prompt revision. An Idea Card fixes the core idea, which is then elaborated into a Research Blueprint. Single-agent and multi-agent workflows share this process.}
\label{fig:zh-2}
\label{fig:architecture}
\floatgap
\end{figure}

\secgap\subsection{InsightPro: An Open Dataset of Insights and Problems}
\label{sec:zh-insightpro}
\label{sec:insightpro}

\begin{figure}[tp]
\centering
\includegraphics[width=\linewidth]{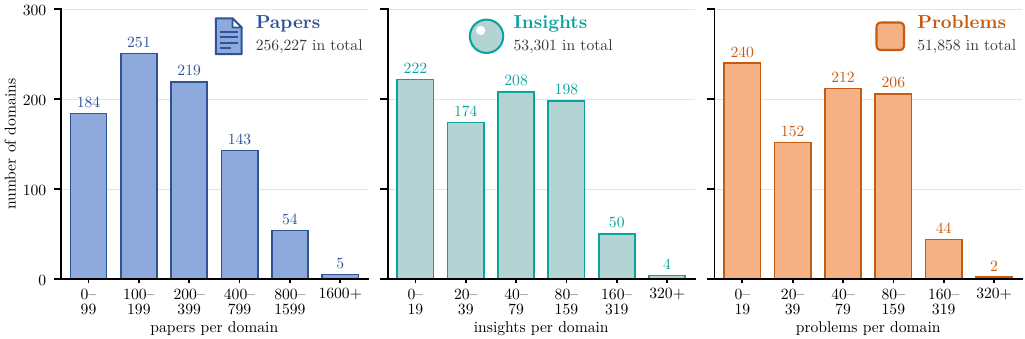}
\figcapgap
\caption{\textbf{Domain distributions in InsightPro.} Numbers of domains in each paper, insight, and problem count range across 856 domains. Papers deduplicated by domain; insight and problem counts include domain- and case-level entries.}
\label{fig:zh-3}
\label{fig:library-statistics}
\floatgap
\end{figure}

\looseness=-1 InsightPro organizes knowledge from \textbf{256,227 papers dated 2016--2026} into \textbf{856 research domains}, containing \textbf{53,301 insights and 51,858 problems}~\citep{einsia2026insightpro}. Sources include major AI venues such as NeurIPS, ICML, ICLR, CVPR, and ACL. Figure~\ref{fig:zh-3} shows the distributions of paper, insight, and problem counts across the 856 domains.

\paragraph{Constructing InsightPro}
InsightPro organizes literature knowledge through four kinds of nodes: domains, papers, insights, and problems. Domains correspond to research topics, such as continual learning or knowledge distillation, bringing related papers and knowledge together. Their construction starts from an initial topic catalog and is refined as papers are assigned. We organize the initial topics using research areas published by major conferences, then summarize papers' core problems and main findings from the available material. Semantic similarity between these descriptions and shared-reference relationships, where available, organize papers into candidate groups. A language model considers these groups alongside existing topics, assessing whether papers in a group address the same core problem and share related core findings, and accordingly adds, merges, or splits topics. Smaller topics are merged with neighboring ones only when their content can support a unified, specific account of problems and insights, yielding 856 domains.

\looseness=-1 Papers are the original studies underlying knowledge entries. Their records retain titles, publication information, and source links, and they are assigned to research topics according to their core problems. For each group, we retrieve candidate topics and use a language model to compare paper content with topic definitions. Changing the method, dataset, or application setting need not create a separate topic; a new topic can be introduced when a group of papers shares a problem not covered by the candidates. Paper memberships are updated as topics are merged or split, producing the final domain--paper links. We also extract structured cards from the available paper material, retaining findings, mechanisms, obstacles, and supporting quotations. The extraction prompt requires unsupported fields to remain empty and restricts abstract-only papers to information explicitly stated in their abstracts.

Insights are concise accounts of prior findings, explaining why a mechanism works and the conditions under which it holds. We extract findings, explanatory mechanisms, conditions of validity, and failure boundaries from the paper cards, then compare them with existing insights in publication order. Findings about the same aspect of the same research object and problem can be organized together, while individual methods' mechanisms and applicability conditions remain within the entry. For opposing conclusions, we examine whether the difference follows from different settings: differences that can be explained yield conditional statements, while unresolved disagreements under comparable conditions retain evidence from both sides.

Problems describe obstacles and unanswered questions in prior research. We extract specific bottlenecks, causes of difficulty, and limitations of existing methods, recording their importance and scope, then progressively merge statements about the same research object and bottleneck within a domain while retaining each paper's description and provenance. Both insights and problems undergo subsequent review to identify missed merges and split entries that mix distinct objects. The number of distinct supporting papers then divides entries into recurring domain-level and less broadly supported case-level knowledge, both retained; this tier reflects breadth of support, not whether a problem is still open, which requires checking later literature.

\looseness=-1 Each insight or problem links to its source papers, allowing readers to begin with a domain overview and trace the original evidence and conditions of validity. A paper may belong to multiple domains and contribute multiple insights or problems. Definitions, construction rules, and counting conventions are given in Appendix~\ref{app:insightpro-details}.

The knowledge-distillation domain illustrates the resulting context reduction: 1,063 papers are represented by 191 insights and 175 problems, whose complete serialized domain export occupies 116,331 tokens. Estimated from a sample of 60 PDFs, the corresponding full texts amount to approximately 15.92M tokens, about 137 times the domain export; the estimation method is described in Appendix~\ref{app:context-budget}.

The \href{https://huggingface.co/datasets/Einsia/InsightPro}{dataset} is publicly released; the \href{https://github.com/Einsia/InsightPro}{code repository} provides the generation prompts and complete examples of using this knowledge for developing research ideas~\citep{einsia2026ideax}.

\secgap\subsection{Research Ideation with InsightPro}
\label{sec:zh-ideax}
\label{sec:zh-blueprint}
\label{sec:ideax}
\label{sec:blueprint}

Building on the InsightPro database, we implement an API for agents to access domains, insights, problems, and source papers, and design prompts for research ideation~\citep{einsia2026ideax}. Agents can browse domains, retrieve relevant entries by keyword, and trace literature evidence through the API, supplementing the information they need with public literature. Users can control the knowledge scope of a generation run through two meta-operations:
\begin{enumerate}[mainlist]
    \item \textbf{Specify the target domain.} Choose the domain for which ideas are generated.
    \item \looseness=-1 \textbf{Recommend domains to cross with.} Suggest domains the agent should connect to, by naming a specific domain in the library or giving a natural-language prompt from which the agent finds suitable domains.
\end{enumerate}
\looseness=-1 The two operations can be combined; without a recommendation, the agent chooses the domains to cross with. Our experiments specify the target domain and leave the choice of cross-domain connections to the agent. Retrieved knowledge supports ideation, and the agent can gather more material as the problem becomes clearer.

\looseness=-1 The prompt first guides the agent to read the available material, understand recurring mechanisms and obstacles in the target domain, and check whether recorded historical problems have been addressed by later work. It then selects at least one core problem worth investigating and explains its importance, the specific difficulty faced by existing methods, and the progress that addressing it could enable. Problems and related insights retain their database sources, and citations record the content obtained from papers for later verification.

Around the core problem, the agent seeks useful insights from the current domain or other domains, explains how this knowledge inspires a new mechanism, and identifies the conditions required for transfer or combination. The prompt provides six exploration perspectives that can be flexibly combined: combining complementary mechanisms and explaining what new capability their interaction enables; transferring across domains through similar objects, structures, or constraints; revisiting the default assumptions of existing methods; distinguishing competing explanations through diagnosis and controls; modifying mechanisms to extend the boundaries of existing capabilities; and exploring freely from a problem or observation. These perspectives can be revisited, and problem selection and mechanism design can be adjusted together as evidence changes.

\looseness=-1 After forming a candidate idea, the agent checks close prior work using the core problem, mechanism operations, and alternative terminology. The prompt requires at least three distinct query formulations and actual reading of relevant sources, comparing the proposal with the mechanisms and conclusions of the closest work. If the core contribution already exists, the agent substantively revises or replaces the candidate; overlap that has not been ruled out remains explicit. This check also distinguishes established literature facts, evidence-based inferences, and new hypotheses requiring validation, so that the remaining contribution and supporting reasoning can be examined. Initial ideation centers on problem value and mechanistic contribution; whether experiments have been completed and the cost of execution do not, on their own, determine an idea's research potential.

Following exploration and revision, the workflow delivers five substantively distinct Idea Cards per domain. Each states the core problem, proposed mechanism, at least three relevant prior works, contribution check, brief validation proposal, failure conditions, and research potential, accompanied by a record of the searches and decisions made. In the single-agent workflow, five isolated sessions per domain each deliver only their top card. The multi-agent workflow starts six independent author agents with the same generation prompt, each producing five cards, followed by one round of discussion and responses over the 30 candidates. Authors may retain, revise, or withdraw their own candidates; an independent selector then chooses five existing cards. Both workflows share the knowledge access API and ideation requirements; the multi-agent workflow adds candidate sampling, discussion, and selection. Table~\ref{tab:system-config} lists the configuration and time limits of every system.

\looseness=-1 A selected card can be further developed into a Research Blueprint explaining why the idea was proposed and how it works. The Blueprint connects the core problem, obstacles revealed by prior work, borrowed insights, transfer assumptions, mechanism design, and testable predictions, adding a problem definition, technical details, concrete examples, comparisons with prior work, and an initial validation plan. The writing process reuses the existing research record, supplements evidence only for factual gaps, citation conflicts, or new claims, and explicitly labels added design choices and unverified assumptions. The original card's core problem, claim, and mechanism remain fixed; if new evidence requires changing the core mechanism, the proposal returns to ideation.

The original card and Blueprint are delivered together, allowing researchers to trace how a core idea is developed and identify assumptions that still require discussion. In our experiments, Blueprints are written from the selected Idea Cards and their research records; their paired evaluation is described in Section~\ref{sec:ablation}. Complete prompts, coordination and writing instructions, and detailed definitions of proposal structure and exploration perspectives are provided in Appendix~\ref{app:prompts}; example ideas in the model efficiency domain are shown on our demo website (Section~\ref{sec:specific}). The experiments use the original Chinese prompts, and the appendix gives their English translation.

\secgap\section{Evaluation}
\label{sec:evaluation}

\begin{figure}[tp]
    \centering
    \includegraphics[width=\linewidth]{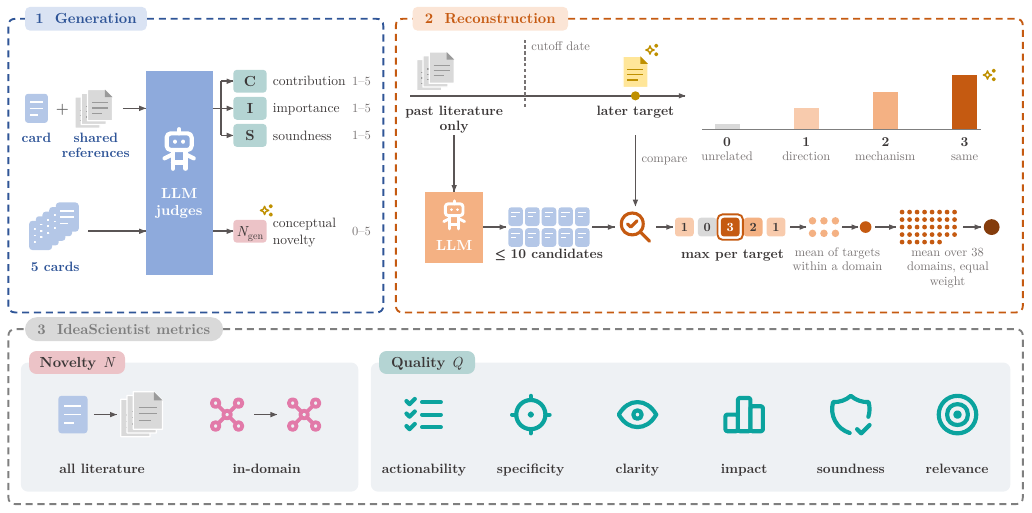}
    \figcapgap
    \caption{\textbf{Evaluation overview.} Generation assesses research value, Reconstruction assesses the ability to rediscover work from earlier literature, and external IdeaScientist metrics evaluate literature novelty and proposal quality.}
    \label{fig:evaluation-method}
\floatgap
\end{figure}

InsightPro's research ideation starts from insights, forms Idea Cards, and then develops selected cards into Research Blueprints.
Insights provide transferable knowledge and mechanisms; an Idea Card states the problem to be solved, the core method, and its expected contribution; a Research Blueprint further specifies technical details, key assumptions, and a validation plan.
The evaluation therefore needs to examine both whether an idea itself has research value and whether it can be developed into a clear, concrete research plan.

We adopt three complementary forms of evaluation, reporting the following metrics:

\begin{itemize}[mainlist]
    \item \textbf{Generation:} judges whether a newly proposed research idea is worth pursuing further, reporting four dimensions, Contribution ($C$), Importance ($I$), Soundness ($S$), and conceptual Novelty ($N$), each scored separately by two independent judge models; see Section~\ref{subsec:evaluation-generation}.
    \item \looseness=-1 \textbf{Reconstruction:} tests whether a system can propose a plan close to the contribution of a later paper using only earlier literature, reporting a 0--3 score with direction and mechanism hit rates; see Section~\ref{subsec:evaluation-reconstruction}.
    \item \looseness=-1 \textbf{External IdeaScientist metrics (IS):} following IdeaScientist, we separately report six proposal-quality dimensions and their mean ($Q$), and literature novelty ($N$); see Section~\ref{subsec:evaluation-ideascientist}.
\end{itemize}

In the main table, both the Generation metrics and the external IdeaScientist metrics take Idea Cards as InsightPro's output, while baseline systems keep their native proposal forms; Section~\ref{sec:ablation} separately compares the same ideas after they are developed from Cards into Blueprints.

\secgap\subsection{Generation: Is the Idea Worth Pursuing?}
\label{subsec:evaluation-generation}
\label{sec:zh-generation}
\label{sec:reviewer}
\label{sec:interesting}

Generation evaluates research ideas that have not yet been carried out, rather than papers with completed experiments.
We evaluate the following four dimensions separately; higher scores indicate better performance on each dimension:

\begin{itemize}[mainlist]
    \item \looseness=-1 \textbf{Contribution ($C$, 1--5):} compared with the closest prior work, what could this idea add? The judge looks for the new understanding, method, or capability it could provide, not merely whether it adopts a new name.
    \item \textbf{Importance ($I$, 1--5):} is the problem worth solving? The judge assesses how important the problem is to its research field and the value that solving it could bring.
    \item \textbf{Soundness ($S$, 1--5):} does the core argument hold? The judge checks whether the key assumptions are clear, whether the method matches the problem, and whether the expected results can reasonably follow from the proposed mechanism. Because the objects of evaluation are research ideas, missing experiments or high implementation costs alone do not warrant a deduction.
    \item \looseness=-1 \textbf{Novelty ($N$, conceptual novelty, 0--5):} does the idea offer an inspiring new perspective? The judge looks for new mechanistic connections, new ways of viewing the problem, and the research directions they open up.
\end{itemize}

\looseness=-1 Each dimension is assessed once by GPT-6 Astra (\texttt{gpt-6-astra}, hereafter Astra) and once by Claude Opus~5.5 (\texttt{claude-opus-5-5}, hereafter Opus~5.5), both with the \texttt{xhigh} reasoning setting.
Each judge assesses every proposal independently once; judges do not know which system generated it and cannot see existing scores or the other judge's results.
The four dimensions are reported separately, and the two judges' results are kept separate.

\looseness=-1 Contribution and conceptual Novelty answer different questions.
Contribution requires comparison with the literature to judge whether the proposal goes beyond existing results; conceptual Novelty examines whether the proposal itself offers a new line of thinking.
An idea can be highly inspiring even though its core mechanism already exists in the literature.

\looseness=-1 Contribution, Importance, and Soundness are scored with reference to literature material.
Based on the proposal's problem and mechanism, we retrieve related papers from a common offline corpus of raw papers and fix the material before scoring; the condensed InsightPro insights and problems are not part of this material.
Both judges receive the same proposal and the same reference material, in which each paper is marked as available in full text, as an abstract, or as metadata only.
Failing to retrieve similar work does not prove that a proposal makes a new contribution.
The organization of the reference material is given in Appendix~\ref{app:generation-scoring}.

\looseness=-1 Conceptual Novelty involves no additional literature retrieval.
Each judge scores every proposal on its own, without seeing the other proposals generated by the same system in the same domain; scoring does not force a ranking or a fixed distribution.

The main table covers 38 domains.
For each judge and each metric, we first compute the mean score within each domain, then average these means equally over domains.
The external IdeaScientist metrics use the same aggregation rule, also scoring Idea Cards; see Section~\ref{subsec:evaluation-ideascientist}.

\secgap\subsection{Reconstruction: Can a System Propose Later Research Contributions?}
\label{subsec:evaluation-reconstruction}
\label{sec:founder}

\looseness=-1 Historical contribution Reconstruction uses published papers as references to test whether a system can propose ideas close to later research from earlier knowledge.
The evaluation set contains 69 target papers across 37 domains, corresponding to 58 tasks.
Each task is jointly defined by a research domain and a literature cutoff date; the number of tasks differs from the number of target papers because one task can correspond to multiple target papers.

The literature cutoff is set to the first day of the target paper's submission month.
The generation stage uses offline, date-constrained literature resources, so that all material explicitly available to the system predates the target paper.
Because the generators' pretraining data may include some target papers, Reconstruction measures whether a system recovers later contributions from date-restricted inputs, not discovery without prior exposure.

For the same task, all systems receive identical task text and submit at most ten candidate proposals for evaluation.
Single runs twice, generating five cards each time.
In Multi, six authors first generate 30 cards, then discuss, revise, and select ten for submission.
IdeaScientist and CoI each run at most ten times, generating one proposal per run.
This limit equalizes the number of evaluated candidates, not the generation cost.
All methods generate these proposals with Opus~4.6 or Opus~4.8.

Astra and Opus~5.5 each compare candidate proposals with the target paper and assign a four-level grade under the same rubric:

\begin{itemize}[mainlist]
    \item \textbf{0:} the proposal is unrelated to the research contribution of the target paper.
    \item \textbf{1:} the research direction is the same, but the mechanism for solving the problem differs.
    \item \textbf{2:} the key mechanism is the same or closely related.
    \item \textbf{3:} both the problem addressed and the mechanism used are substantially the same.
\end{itemize}

System outputs are anonymized before evaluation, and their presentation order is determined by a seeded shuffle.
For each target paper, we take the highest grade among up to ten candidates as the system's Reconstruction score for that paper.
A best grade of at least 1 counts as a direction hit; a best grade of at least 2 counts as a mechanism hit.
These metrics measure whether an idea close to the target contribution appears in the candidate set, not whether a selector can rank it first.

For aggregation, we first average the Reconstruction scores of the target papers within each domain, then average equally over the 37 domains; the final score remains on the 0--3 scale.

\secgap\subsection{External IdeaScientist Metrics: Novelty and Proposal Quality}
\label{subsec:evaluation-ideascientist}
\label{sec:isbench}

We follow IdeaScientist's public evaluation method~\citep{liu2026ideascientist} to further judge how new a proposal is relative to the literature and whether its research plan is clear, concrete, and executable.
This subsection explains the metrics we use; the metric set, scoring scales, and differences from the original method adopted in this paper are detailed in Appendix~\ref{app:is-metrics-adaptation}.

\paragraph{Original IdeaScientist metrics.}
IdeaScientist organizes its evaluation into two groups, Novelty ($N$, distinct from the $N$ metric of idea Generation in Section~\ref{subsec:evaluation-generation}) and proposal Quality ($Q$), and reports an Overall score.
Of the Novelty metrics, we use \textbf{All}, which uses related papers retrieved from the complete comparison corpus to judge which existing mechanisms the proposal goes beyond.
Quality is the equally weighted mean of six quality dimensions; All and the quality dimensions are scored on a 1--10 scale.

\paragraph{Proposal quality.}
The six dimensions examine different aspects of the research plan:

\begin{itemize}[mainlist]
    \item \textbf{Actionability:} could a researcher start implementing the plan from it? Are the method steps, the key design choices, and the validation plan sufficiently explicit?
    \item \textbf{Specificity:} does the plan give substantive technical details, such as mechanisms, formulas, datasets, baselines, or metrics, rather than only general research goals?
    \item \looseness=-1 \textbf{Clarity:} are the problem, claims, and method clearly organized, and is the writing precise and easy to read?
    \item \textbf{Impact:} if the plan succeeds, how much could it contribute to the research field? This judgment concerns expected value and does not treat experimental results not yet obtained as facts.
    \item \textbf{Soundness:} is the technical reasoning credible and internally consistent? Are the key assumptions explicit, do the arguments support the claims, and are possible failure cases acknowledged?
    \item \textbf{Relevance:} does the plan directly address the problem and difficulty it sets out to solve, or does it drift from the goal, handle only part of it, or give a generic answer?
\end{itemize}

\looseness=-1 Actionability concerns whether the plan can be executed, and Specificity concerns whether its details are sufficient; the two differ: a detailed plan may still lack key implementable steps.
Soundness concerns whether the argument holds, whereas Impact concerns whether the plan would be valuable if it held.
These dimensions are judged from what the proposal explicitly states, without filling in unstated designs or evidence for the authors.

\paragraph{Literature comparison in Novelty (All).}
This metric answers the question: relative to the retrieved related work, what exactly does the proposal's core method add?
We first retrieve related papers, then organize the reference material by their research problems, main difficulties, and solution mechanisms, and fix the material before scoring.
Both judges receive the same reference content and compare the overlap and differences between the proposal and prior work.
Unlike the assessment of Contribution in Section~\ref{subsec:evaluation-generation}, this judgment focuses on mechanism-level overlap and differences, so the comparison material uses only the structured content above and excludes paper abstracts and empirical outcome passages.

Following IdeaScientist's original scoring rationale, restating existing mechanisms, renaming them, or directly concatenating existing methods should receive low ratings; only a proposal that clearly identifies a key mechanism or substantive change distinct from the reference work has grounds for a higher rating.
It differs from the conceptual Novelty of Section~\ref{subsec:evaluation-generation}: the latter considers only whether the idea is inspiring and involves no additional literature comparison.

\secgap\section{Experiments}
\subsection{Experimental Setup}

We design ideation prompts based on the InsightPro database and compare their single-agent (Single) and multi-agent (Multi) execution with nine baseline systems: IdeaScientist, CoI, AutoResearchClaw, Dr.~Claw, DeepScientist, ResearchAgent, SciAgents, CycleResearcher, and the native idea stage of AI Scientist v3; raw-paper retrieval (RAG) serves as an ablation reference (Section~\ref{sec:ablation}).

Every baseline is adapted from its official implementation and keeps its native workflow with the shared generation model; Table~\ref{tab:system-config} in Appendix~\ref{app:experiments} lists the source, native workflow, literature access, and adaptations of every system.
Scores are averaged over the successfully generated ideas: outputs where a system terminates without delivering a final idea (native failures, Table~\ref{tab:native-failures}) are excluded, and infrastructure and adaptation failures are rerun.
Open-ended generation covers the same 38 domains with five output slots per domain and uses five generation models: GPT-6 Astra (xhigh), Qwen3.8-27B, Qwen3.8-Flash-Next, DeepSeek-V4.1-Flash-1017, and GLM-5.3-Flash; Qwen models use a total context of 262,144 tokens per request.
\looseness=-1 We evaluate the baselines' native proposals and pair the Idea Cards generated with our prompts with their Research Blueprints to study proposal development.

\looseness=-1 Astra/xhigh and Opus~5.5/xhigh each assess every output once, and scores are aggregated as in Section~\ref{subsec:evaluation-generation}.

Reconstruction uses offline, date-constrained historical adaptations, evaluated on 69 targets in 58 tasks across 37 domains.
Every method generates up to ten candidates per task.
Proposal generation uses the date-selected Opus~4.6/4.8 configuration, and Astra and Opus~5.5 independently judge the degree of reconstruction.
We take the highest grade among the candidates for each target, compute domain means, and average equally over the 37 domains; grades remain on the 0--3 scale, and direction and mechanism hits are the shares of targets whose best grade is at least 1 and at least 2, respectively.

\secgap\subsection{Main Results}
\begingroup
\tabcaptop
\footnotesize
\setlength{\tabcolsep}{9.87pt}
\setlength{\LTleft}{\fill}
\setlength{\LTright}{\fill}
\setlength{\LTpost}{0pt}
\renewcommand{\arraystretch}{1.0}
\begin{longtable}{@{}lcccccc@{}}
\caption{\textbf{Main results by generation model.} Each block lists the 11 systems for one generation model; each numeric cell is \textbf{Astra / Opus} (Astra: gpt-6-astra; Opus: Claude Opus~5.5; both xhigh), the two judges reported separately. Reconstruction uses Claude Opus~4.6 or Opus~4.8 for all methods and is reported per method in Table~\ref{tab:recon}. Results of our workflows with the Claude Opus generator are in Table~\ref{tab:claude-opus} in the appendix.}
\label{tab:zh-3}\label{tab:main}\\
\toprule
 & \multicolumn{4}{c}{\textbf{Generation}} & \multicolumn{2}{c}{\textbf{IS}} \\
\cmidrule(lr){2-5}\cmidrule(lr){6-7}
Method & Contribution & Importance & Soundness & Novelty & Lit.\ novelty & Quality \\
 & (1--5) & (1--5) & (1--5) & (0--5) & (0--100) & (0--100) \\
\midrule
\endfirsthead
\multicolumn{7}{@{}l}{\textit{Table continued}}\\
\toprule
 & \multicolumn{4}{c}{\textbf{Generation}} & \multicolumn{2}{c}{\textbf{IS}} \\
\cmidrule(lr){2-5}\cmidrule(lr){6-7}
Method & Contribution & Importance & Soundness & Novelty & Lit.\ novelty & Quality \\
 & (1--5) & (1--5) & (1--5) & (0--5) & (0--100) & (0--100) \\
\midrule
\endhead
\midrule\multicolumn{7}{r@{}}{\textit{Continued on next page}}\\
\endfoot
\bottomrule
\endlastfoot
\multicolumn{7}{@{}l}{\textbf{GPT-6 Astra}}\\*
\hspace{0.4em}IdeaScientist & 3.01\,/\,2.66 & 3.55\,/\,2.99 & 3.75\,/\,3.37 & 3.76\,/\,3.08 & 53.80\,/\,52.40 & 79.30\,/\,68.40 \\
\hspace{0.4em}CoI & 3.06\,/\,2.83 & 3.46\,/\,3.11 & 3.82\,/\,3.48 & 3.84\,/\,3.11 & 52.80\,/\,50.40 & 80.00\,/\,70.40 \\
\hspace{0.4em}AutoResearchClaw & 3.02\,/\,2.81 & 3.38\,/\,3.02 & 3.46\,/\,3.14 & 3.71\,/\,3.21 & 56.50\,/\,54.40 & 71.80\,/\,63.40 \\
\hspace{0.4em}Dr.~Claw & 2.98\,/\,2.52 & 3.16\,/\,3.02 & 3.88\,/\,3.26 & 3.51\,/\,2.99 & 47.20\,/\,45.50 & 80.80\,/\,72.50 \\
\hspace{0.4em}DeepScientist & 3.00\,/\,2.77 & 3.38\,/\,3.09 & 3.93\,/\,3.48 & 3.55\,/\,3.13 & 47.70\,/\,48.30 & {\rmfamily\bfseries 82.80}\,/\,75.10 \\
\hspace{0.4em}ResearchAgent & 3.02\,/\,2.83 & 3.61\,/\,2.97 & {\rmfamily\bfseries 4.01}\,/\,3.48 & 3.76\,/\,3.17 & 58.50\,/\,55.30 & 81.20\,/\,76.30 \\
\hspace{0.4em}SciAgents & 2.77\,/\,2.26 & 2.94\,/\,2.63 & 3.53\,/\,3.11 & 3.08\,/\,2.60 & 41.40\,/\,41.40 & 73.50\,/\,71.10 \\
\hspace{0.4em}CycleResearcher & 2.98\,/\,2.65 & 3.56\,/\,3.07 & 3.82\,/\,3.37 & 3.76\,/\,3.07 & 52.10\,/\,50.20 & 80.20\,/\,72.90 \\
\hspace{0.4em}AI Scientist v3 & 3.06\,/\,2.52 & 3.64\,/\,3.05 & 3.38\,/\,3.19 & 3.70\,/\,2.93 & 50.20\,/\,47.90 & 78.30\,/\,67.20 \\
\rowcolor{brandbg}\hspace{0.4em}Single (ours) & 3.35\,/\,{\rmfamily\bfseries 3.02} & 4.12\,/\,3.24 & 3.58\,/\,3.47 & 3.92\,/\,3.48 & 57.74\,/\,55.26 & 81.40\,/\,75.00 \\*
\rowcolor{brandbg}\hspace{0.4em}Multi (ours) & {\rmfamily\bfseries 3.41}\,/\,{\rmfamily\bfseries 3.02} & {\rmfamily\bfseries 4.18}\,/\,{\rmfamily\bfseries 3.26} & 3.93\,/\,{\rmfamily\bfseries 3.65} & {\rmfamily\bfseries 4.00}\,/\,{\rmfamily\bfseries 3.55} & {\rmfamily\bfseries 59.89}\,/\,{\rmfamily\bfseries 57.37} & 82.10\,/\,{\rmfamily\bfseries 76.70} \\
\midrule\multicolumn{7}{@{}l}{\textbf{Qwen3.8-27B}}\\*
\hspace{0.4em}IdeaScientist & 2.75\,/\,2.29 & 3.25\,/\,2.50 & 2.50\,/\,2.62 & 2.39\,/\,2.30 & 38.60\,/\,43.60 & 65.10\,/\,62.80 \\
\hspace{0.4em}CoI & 2.86\,/\,2.27 & 3.07\,/\,2.46 & {\rmfamily\bfseries 2.77}\,/\,2.89 & 2.99\,/\,2.28 & 41.50\,/\,46.10 & {\rmfamily\bfseries 71.80}\,/\,67.70 \\
\hspace{0.4em}AutoResearchClaw & 2.86\,/\,2.37 & 2.79\,/\,2.35 & 2.53\,/\,2.59 & 3.03\,/\,2.26 & 40.10\,/\,39.30 & 59.00\,/\,54.10 \\
\hspace{0.4em}Dr.~Claw & 2.67\,/\,2.08 & 2.64\,/\,2.19 & 2.53\,/\,2.50 & 2.44\,/\,2.16 & 39.30\,/\,42.10 & 67.30\,/\,64.50 \\
\hspace{0.4em}DeepScientist & 2.90\,/\,2.49 & 3.13\,/\,2.59 & 2.62\,/\,2.87 & 2.85\,/\,2.43 & 40.40\,/\,42.70 & 69.40\,/\,65.60 \\
\hspace{0.4em}ResearchAgent & 2.95\,/\,2.53 & 2.83\,/\,2.45 & 2.54\,/\,{\rmfamily\bfseries 2.92} & {\rmfamily\bfseries 3.04}\,/\,2.35 & {\rmfamily\bfseries 44.70}\,/\,44.10 & 70.30\,/\,{\rmfamily\bfseries 70.40} \\
\hspace{0.4em}SciAgents & 2.93\,/\,2.61 & 2.77\,/\,2.47 & 2.56\,/\,2.69 & 2.50\,/\,2.49 & 41.40\,/\,44.60 & 59.00\,/\,63.60 \\
\hspace{0.4em}CycleResearcher & 2.45\,/\,2.12 & 2.66\,/\,2.24 & 2.58\,/\,2.25 & 2.30\,/\,2.03 & 34.50\,/\,36.10 & 63.10\,/\,56.30 \\
\hspace{0.4em}AI Scientist v3 & 2.72\,/\,2.13 & 2.67\,/\,2.33 & 2.69\,/\,2.71 & 2.75\,/\,2.01 & 40.40\,/\,37.70 & 69.70\,/\,62.50 \\
\rowcolor{brandbg}\hspace{0.4em}Single (ours) & 2.95\,/\,{\rmfamily\bfseries 2.76} & 3.32\,/\,2.67 & 2.71\,/\,2.79 & 2.83\,/\,{\rmfamily\bfseries 2.77} & 43.00\,/\,46.10 & 70.20\,/\,68.90 \\*
\rowcolor{brandbg}\hspace{0.4em}Multi (ours) & {\rmfamily\bfseries 3.00}\,/\,2.75 & {\rmfamily\bfseries 3.37}\,/\,{\rmfamily\bfseries 2.73} & 2.74\,/\,2.80 & 2.95\,/\,2.76 & 44.30\,/\,{\rmfamily\bfseries 48.60} & 70.90\,/\,69.50 \\
\midrule\multicolumn{7}{@{}l}{\textbf{DeepSeek-V4.1-Flash-1017}}\\*
\hspace{0.4em}IdeaScientist & 2.99\,/\,2.85 & 3.42\,/\,2.81 & 3.21\,/\,2.92 & 2.77\,/\,3.28 & 44.90\,/\,48.20 & 66.40\,/\,64.10 \\
\hspace{0.4em}CoI & 2.92\,/\,2.48 & 3.30\,/\,2.81 & 3.48\,/\,3.36 & 3.25\,/\,2.71 & 47.40\,/\,51.10 & 70.10\,/\,68.70 \\
\hspace{0.4em}AutoResearchClaw & 2.99\,/\,2.79 & 3.33\,/\,2.81 & 2.98\,/\,3.10 & 3.46\,/\,3.04 & 48.60\,/\,49.30 & 62.30\,/\,59.20 \\
\hspace{0.4em}Dr.~Claw & 3.01\,/\,2.79 & 3.45\,/\,2.90 & 3.25\,/\,3.02 & 3.00\,/\,3.25 & 43.10\,/\,46.40 & 69.20\,/\,66.10 \\
\hspace{0.4em}DeepScientist & 3.00\,/\,2.76 & 3.48\,/\,2.99 & 3.36\,/\,3.16 & 3.19\,/\,3.13 & 44.20\,/\,45.80 & 69.80\,/\,65.80 \\
\hspace{0.4em}ResearchAgent & 2.98\,/\,2.75 & 3.36\,/\,2.76 & 3.13\,/\,3.38 & 3.44\,/\,2.85 & 47.80\,/\,50.30 & 70.60\,/\,69.30 \\
\hspace{0.4em}SciAgents & 2.94\,/\,2.62 & 2.90\,/\,2.57 & 3.19\,/\,2.98 & 2.70\,/\,3.00 & 42.60\,/\,45.70 & 61.70\,/\,62.90 \\
\hspace{0.4em}CycleResearcher & 2.83\,/\,2.32 & 3.21\,/\,2.75 & 3.28\,/\,3.21 & 3.12\,/\,2.83 & 41.80\,/\,43.90 & 65.20\,/\,60.40 \\
\hspace{0.4em}AI Scientist v3 & 2.92\,/\,2.44 & 3.34\,/\,2.84 & 3.39\,/\,3.28 & 3.29\,/\,2.93 & 41.50\,/\,40.20 & 68.50\,/\,61.80 \\
\rowcolor{brandbg}\hspace{0.4em}Single (ours) & 3.02\,/\,2.99 & 3.90\,/\,3.20 & 3.50\,/\,3.41 & 3.65\,/\,3.32 & 49.10\,/\,53.20 & 70.20\,/\,68.90 \\*
\rowcolor{brandbg}\hspace{0.4em}Multi (ours) & {\rmfamily\bfseries 3.10}\,/\,{\rmfamily\bfseries 3.01} & {\rmfamily\bfseries 3.98}\,/\,{\rmfamily\bfseries 3.24} & {\rmfamily\bfseries 3.56}\,/\,{\rmfamily\bfseries 3.46} & {\rmfamily\bfseries 3.77}\,/\,{\rmfamily\bfseries 3.38} & {\rmfamily\bfseries 49.50}\,/\,{\rmfamily\bfseries 53.80} & {\rmfamily\bfseries 70.80}\,/\,{\rmfamily\bfseries 69.60} \\
\midrule\multicolumn{7}{@{}l}{\textbf{GLM-5.3-Flash}}\\*
\hspace{0.4em}IdeaScientist & 2.87\,/\,2.60 & 3.10\,/\,2.68 & 3.19\,/\,3.16 & 3.15\,/\,2.99 & 46.30\,/\,49.20 & 66.20\,/\,64.60 \\
\hspace{0.4em}CoI & 3.04\,/\,2.74 & 3.69\,/\,3.05 & 3.35\,/\,3.28 & 3.39\,/\,3.04 & 47.40\,/\,51.50 & 70.40\,/\,69.70 \\
\hspace{0.4em}AutoResearchClaw & 3.08\,/\,2.94 & 3.79\,/\,3.10 & 3.07\,/\,3.33 & 3.75\,/\,3.15 & {\rmfamily\bfseries 49.30}\,/\,46.80 & 61.10\,/\,57.90 \\
\hspace{0.4em}Dr.~Claw & 2.98\,/\,2.57 & 3.58\,/\,3.02 & 3.31\,/\,3.12 & 3.11\,/\,2.98 & 42.00\,/\,45.20 & 69.40\,/\,67.10 \\
\hspace{0.4em}DeepScientist & 2.85\,/\,2.47 & 3.47\,/\,2.93 & {\rmfamily\bfseries 3.57}\,/\,3.29 & 3.33\,/\,2.82 & 42.20\,/\,43.70 & 68.80\,/\,62.00 \\
\hspace{0.4em}ResearchAgent & 3.03\,/\,2.88 & 3.69\,/\,3.11 & 3.34\,/\,{\rmfamily\bfseries 3.40} & 3.53\,/\,3.24 & 47.50\,/\,48.50 & 70.30\,/\,{\rmfamily\bfseries 73.00} \\
\hspace{0.4em}SciAgents & 2.71\,/\,2.62 & 3.09\,/\,2.64 & 3.15\,/\,2.99 & 2.86\,/\,2.97 & 45.30\,/\,47.80 & 60.20\,/\,67.80 \\
\hspace{0.4em}CycleResearcher & 2.74\,/\,2.48 & 2.88\,/\,2.43 & 3.14\,/\,3.09 & 3.01\,/\,2.75 & 44.90\,/\,46.50 & 65.80\,/\,62.90 \\
\hspace{0.4em}AI Scientist v3 & 2.53\,/\,2.10 & 2.95\,/\,2.63 & 3.39\,/\,3.08 & 2.95\,/\,2.28 & 38.90\,/\,37.40 & 69.00\,/\,58.00 \\
\rowcolor{brandbg}\hspace{0.4em}Single (ours) & 3.07\,/\,2.94 & 3.83\,/\,3.18 & 3.27\,/\,3.32 & 3.82\,/\,3.31 & 48.00\,/\,53.40 & 71.00\,/\,71.50 \\*
\rowcolor{brandbg}\hspace{0.4em}Multi (ours) & {\rmfamily\bfseries 3.13}\,/\,{\rmfamily\bfseries 3.00} & {\rmfamily\bfseries 3.91}\,/\,{\rmfamily\bfseries 3.23} & 3.50\,/\,{\rmfamily\bfseries 3.40} & {\rmfamily\bfseries 3.90}\,/\,{\rmfamily\bfseries 3.37} & 48.60\,/\,{\rmfamily\bfseries 53.50} & {\rmfamily\bfseries 71.60}\,/\,72.20 \\
\midrule\multicolumn{7}{@{}l}{\textbf{Qwen3.8-Flash-Next}}\\*
\hspace{0.4em}IdeaScientist & 2.94\,/\,2.82 & 3.15\,/\,2.55 & 2.99\,/\,2.95 & 2.81\,/\,2.90 & 47.30\,/\,50.10 & 64.00\,/\,63.90 \\
\hspace{0.4em}CoI & 2.85\,/\,2.39 & 3.25\,/\,2.64 & {\rmfamily\bfseries 3.34}\,/\,{\rmfamily\bfseries 3.22} & 3.07\,/\,2.20 & 42.60\,/\,46.30 & 72.60\,/\,67.90 \\
\hspace{0.4em}AutoResearchClaw & 2.94\,/\,2.46 & 3.17\,/\,2.57 & 3.08\,/\,2.93 & {\rmfamily\bfseries 3.13}\,/\,2.30 & 43.60\,/\,44.80 & 60.50\,/\,54.90 \\
\hspace{0.4em}Dr.~Claw & 2.89\,/\,2.71 & 3.15\,/\,2.60 & 3.06\,/\,2.95 & 2.88\,/\,2.75 & 47.10\,/\,49.80 & 68.90\,/\,68.00 \\
\hspace{0.4em}DeepScientist & 2.87\,/\,2.76 & 3.21\,/\,2.64 & 3.13\,/\,3.00 & 2.85\,/\,2.73 & 46.60\,/\,49.90 & 69.40\,/\,67.70 \\
\hspace{0.4em}ResearchAgent & 2.95\,/\,2.63 & 3.17\,/\,2.60 & 3.08\,/\,3.20 & 3.09\,/\,2.28 & 45.50\,/\,44.90 & 71.20\,/\,70.30 \\
\hspace{0.4em}SciAgents & 2.61\,/\,2.33 & 2.49\,/\,2.21 & 2.96\,/\,2.93 & 2.49\,/\,2.09 & 41.90\,/\,42.30 & 57.90\,/\,61.30 \\
\hspace{0.4em}CycleResearcher & 2.28\,/\,2.18 & 2.70\,/\,2.36 & 3.01\,/\,2.87 & 2.35\,/\,2.08 & 36.60\,/\,35.30 & 66.00\,/\,58.70 \\
\hspace{0.4em}AI Scientist v3 & 2.91\,/\,2.50 & 3.05\,/\,2.51 & 2.96\,/\,2.83 & 3.01\,/\,2.38 & 47.00\,/\,45.50 & 72.10\,/\,66.90 \\
\rowcolor{brandbg}\hspace{0.4em}Single (ours) & 2.96\,/\,2.78 & 3.44\,/\,2.75 & 3.26\,/\,3.12 & 2.96\,/\,2.95 & 47.80\,/\,50.60 & {\rmfamily\bfseries 72.90}\,/\,69.60 \\*
\rowcolor{brandbg}\hspace{0.4em}Multi (ours) & {\rmfamily\bfseries 3.00}\,/\,{\rmfamily\bfseries 2.90} & {\rmfamily\bfseries 3.50}\,/\,{\rmfamily\bfseries 2.82} & 3.30\,/\,3.09 & 3.04\,/\,{\rmfamily\bfseries 3.10} & {\rmfamily\bfseries 48.30}\,/\,{\rmfamily\bfseries 51.30} & 71.70\,/\,{\rmfamily\bfseries 70.70} \\
\end{longtable}
\endgroup

\begin{table}[H]
\tabcaptop
\centering
\footnotesize
\caption{\textbf{Reconstruction results by method.} For all methods, candidates are generated by Claude Opus~4.6 or Opus~4.8 and evaluated on 69 targets in 58 tasks across 37 domains. Each numeric cell is \textbf{Astra / Opus} (judges). Each task has at most ten candidate cards; Grade is the per-target best grade among the candidates (0--3), and direction and mechanism hits are the shares of targets whose best grade is at least 1 and at least 2; all are averaged within each domain and then equally over the 37 domains, with native failures counted as zero. Bold marks the best value per column and judge.}
\label{tab:recon}
\setlength{\tabcolsep}{6.00pt}
\renewcommand{\arraystretch}{1.0}
\begin{tabular}{@{}lccc@{}}
\toprule
Method & Grade (0--3) $\uparrow$ & Direction hit $\uparrow$ & Mechanism hit $\uparrow$ \\
\midrule
IdeaScientist & 0.76 / 0.43 & 67\% / 40\% & 10\% / 4\% \\
CoI & 0.57 / 0.41 & 55\% / 38\% & 3\% / 4\% \\
AutoResearchClaw & 0.73 / 0.51 & 63\% / 48\% & 10\% / 3\% \\
Dr.~Claw & 0.58 / 0.41 & 49\% / 37\% & 9\% / 4\% \\
DeepScientist & 0.64 / 0.47 & 53\% / 40\% & 11\% / 6\% \\
ResearchAgent & 0.52 / 0.34 & 46\% / 30\% & 7\% / 4\% \\
SciAgents & 0.77 / 0.49 & 70\% / 45\% & 10\% / 5\% \\
CycleResearcher & 0.53 / 0.33 & 43\% / 33\% & 10\% / 1\% \\
AI Scientist v3 & 0.50 / 0.38 & 47\% / 38\% & 4\% / 0\% \\
\rowcolor{brandbg}Single (ours) & \textbf{0.94} / 0.73 & 75\% / 66\% & 19\% / 7\% \\
\rowcolor{brandbg}Multi (ours) & 0.93 / \textbf{0.75} & \textbf{79\%} / \textbf{68\%} & \textbf{21\%} / \textbf{11\%} \\
\midrule
Raw-paper RAG & 0.62 / 0.44 & 48\% / 36\% & 11\% / 8\% \\
\bottomrule
\end{tabular}
\floatgap
\end{table}

Table~\ref{tab:zh-3} evaluates the 11 systems under five generation models.
Our workflows lead in contribution, importance, novelty, and literature novelty.
Averaged over the two judges, Multi scores above the strongest baseline on all four with every generation model.
Averaged over the five generation models as well, it leads the strongest baseline by 10.7\% in importance (3.42 vs.\ 3.09) and 9.0\% in novelty (3.38 vs.\ 3.10), with smaller margins in contribution and literature novelty.
Single follows closely and ranks second on all four.

Our advantage comes mainly from these dimensions, which measure the value of an idea.
In soundness and proposal quality, our workflows are in the first tier, on par with the strongest baseline: on average, Multi is slightly above CoI in soundness and ResearchAgent in quality, and the per-model differences go both ways.
Section~\ref{sec:ablation} shows that both dimensions mainly track how detailed the description is, and that both rise once the concise Idea Cards are elaborated into Research Blueprints.

Scores also follow the strength of the generation model: GPT-6 Astra $>$ GLM-5.3-Flash $\approx$ DeepSeek-V4.1-Flash-1017 $\geq$ Qwen3.8-Flash-Next $>$ Qwen3.8-27B.
Averaged over all systems, metrics, and judges, the Generation score falls from 3.28 with GPT-6 Astra to 2.63 with Qwen3.8-27B, and the ordering holds under both judges and for nearly all systems.
Our workflows lead under every generation model, so their gains hold across generation models of different strength.

Because all methods use the same generators for Reconstruction, Table~\ref{tab:recon} reports it once per method.
Our workflows recover historical contributions best: Single and Multi reach the highest grades under both judges, about a third above the strongest baseline on average (0.84 vs.\ 0.63 for SciAgents), and Multi nearly doubles the mechanism-hit share of the strongest baseline, DeepScientist.

\secgap\subsection{Comparison under IdeaScientist Metrics}
\label{sec:is-outputs}
The external IdeaScientist rubric reports literature novelty $N$ and proposal quality $Q$, the equal-weight mean of six quality dimensions (Section~\ref{subsec:evaluation-ideascientist}); all systems are scored on Idea Cards or native proposals.
These metrics agree with the Generation results.
Averaged over the two judges, Multi has the highest literature novelty with every generation model, leading the strongest baseline by 2.3\% to 4.9\%.
Its quality is on par with the strongest baseline: the gap is within 1\% with every generation model, and Multi's average is slightly higher than that of ResearchAgent.
Idea Cards are deliberately concise, while actionability, specificity, and related dimensions mainly track how detailed a description is; once the same Cards are elaborated into Research Blueprints, the quality of Multi with Astra rises from 79.4 to 82.2 (Section~\ref{sec:ablation}).

\secgap\subsection{Agreement with Human Experts}
\label{sec:ai-review}
\begin{figure}[tp]
\centering
\includegraphics[width=\linewidth]{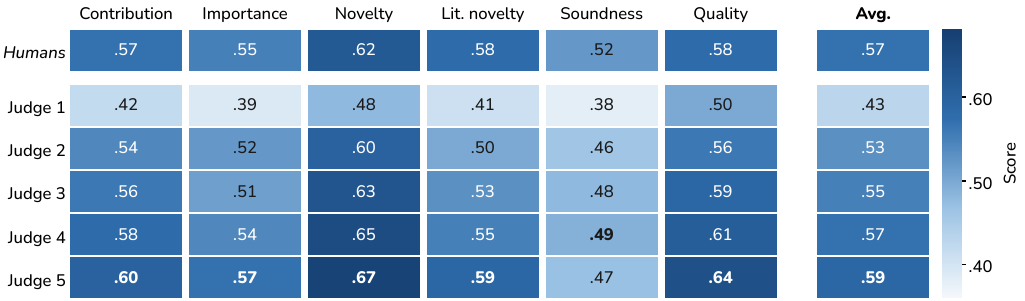}
\caption{\textbf{Agreement between LLM judges and human experts.} Each cell is the item-level Pearson correlation between a judge's scores and the human scores on one dimension; the first row is the correlation between human experts, and Avg.\ averages the six dimensions. Judges~1--5 are Qwen3.8-27B, Qwen3.8-Flash-Next, GLM-5.3-Flash, DeepSeek-V4.1-Flash-1017, and GPT-6 Astra; bold marks the highest judge correlation in each column.}
\label{fig:judge-agreement}
\floatgap
\end{figure}

To check whether the LLM judges agree with human judgment, three human experts scored a blinded sample of ideas, and we compare their scores with those of LLM judges of different strength (Figure~\ref{fig:judge-agreement}).
The sample uses the Astra generator and covers 12 domains, excluding those used in prompt development, and six systems: Single, Multi, and the four baselines with the highest mean Generation scores from the two judges (ResearchAgent, CoI, DeepScientist, and CycleResearcher).
One idea per system and domain was drawn with a fixed seed, giving 72 ideas.
Each idea was rendered as readable text with system names, knowledge-base identifiers, and internal links removed, but it was not rewritten into a common format; reviewers saw only an item ID, the domain task text, and the idea.
The experts scored the ideas on the six dimensions used in our evaluation: Contribution, Importance, Novelty, literature novelty, Soundness, and quality.

Averaged over the six dimensions, the correlation between human experts is 0.57.
GPT-6 Astra, a judge in our evaluation, correlates with the experts at 0.59, as closely as the experts agree with one another, and it does so on every dimension except Soundness.
Agreement grows with the strength of the judge, from 0.43 for Qwen3.8-27B upward, in the same ordering as for generation.
Soundness is the hardest dimension to agree on, for the experts as well as for every judge.
On this sample, GPT-6 Astra agrees with the experts about as well as the experts agree with one another, which supports using it as a judge.

\secgap\subsection{Ablations}
\label{sec:ablation}

This section reports two comparisons: the same ideas in Idea Card and Research Blueprint form, and our workflows against a raw-paper retrieval reference (RAG).

\begin{table}[tp]
\footnotesize
\tabcaptop
\centering
\footnotesize
\caption{\textbf{Paired Idea Cards and Research Blueprints (Astra generator).} Each numeric cell is \textbf{Astra / Opus} (judges); each Blueprint elaborates the same Card. As in the Astra block of the main table, domain means are computed first (for quality, per dimension) and the 38 domains are then equal-weighted. Evidence was retrieved independently for each form at scoring. Contribution, Importance, and Soundness use the same scoring as the main table; literature novelty and quality are IS metrics. Bold marks the higher value within a system.}
\label{tab:card-blueprint}
\setlength{\tabcolsep}{12.63pt}
\begin{tabular}{@{}llccccc@{}}
\toprule
Workflow & Output & Contribution & Importance & Soundness & Lit.\ novelty & Quality \\
 & & (1--5) & (1--5) & (1--5) & (0--100) & (0--100) \\
\midrule
Single (ours) & Card & {\rmfamily\bfseries 3.35}\,/\,3.02 & {\rmfamily\bfseries 4.12}\,/\,{\rmfamily\bfseries 3.24} & 3.58\,/\,3.47 & 57.74\,/\,55.26 & 81.40\,/\,75.00 \\
 & Blueprint & 3.26\,/\,{\rmfamily\bfseries 3.03} & 3.96\,/\,3.09 & {\rmfamily\bfseries 4.12}\,/\,{\rmfamily\bfseries 3.80} & {\rmfamily\bfseries 60.74}\,/\,{\rmfamily\bfseries 57.42} & {\rmfamily\bfseries 84.97}\,/\,{\rmfamily\bfseries 77.92} \\
\midrule
Multi (ours) & Card & {\rmfamily\bfseries 3.41}\,/\,3.02 & {\rmfamily\bfseries 4.18}\,/\,{\rmfamily\bfseries 3.26} & 3.93\,/\,3.65 & 59.89\,/\,57.37 & 82.10\,/\,76.70 \\
 & Blueprint & 3.27\,/\,{\rmfamily\bfseries 3.06} & 3.95\,/\,3.09 & {\rmfamily\bfseries 4.28}\,/\,{\rmfamily\bfseries 3.88} & {\rmfamily\bfseries 63.11}\,/\,{\rmfamily\bfseries 59.47} & {\rmfamily\bfseries 85.23}\,/\,{\rmfamily\bfseries 79.26} \\
\bottomrule
\end{tabular}
\floatgap
\end{table}

\paragraph{Idea Cards and Research Blueprints.}

\looseness=-1 This comparison asks what Soundness measures.
Blueprints develop the selected Cards: the core idea stays fixed, and research records, explicit mechanism details, and targeted evidence acquisition only make the technical argument more complete (Section~\ref{sec:zh-blueprint}).
A dimension that changes markedly between a Card and its Blueprint therefore reflects how detailed the description is rather than the idea itself.
Table~\ref{tab:card-blueprint} uses the Astra generator and the same aggregation as the Astra block of the main table.

The two kinds of dimensions diverge clearly.
Dimensions that measure the value of the idea do not rise: Contribution changes by at most 0.14 points, and Importance drops slightly.
Soundness, which measures whether the argument is complete, rises consistently for both workflows, for Multi from 3.79 to 4.08 averaged over the judges.

Literature novelty also rises, as a fuller description makes differences from prior work easier for judges to identify, and quality rises as well.
For the same idea, Soundness and quality therefore depend mainly on how detailed the description is and have little to do with the quality of the idea itself.
Blueprint writing allows further retrieval and scoring retrieves evidence anew, so the differences include the whole elaboration stage.

\paragraph{Comparison with raw-paper retrieval.}
\begin{table}[tp]
\footnotesize
\tabcaptop
\centering
\caption{\textbf{Comparison with raw-paper retrieval.} Each numeric cell is \textbf{Astra / Opus} (judges); metrics and aggregation follow Table~\ref{tab:main}. RAG retrieves papers from the same sources as InsightPro and generates one idea from them, without access to InsightPro's insights and problems; all rows are scored as Idea Cards. The average weights the four generation models equally. Bold marks the best value per block, column and judge.}
\label{tab:ablation}
\setlength{\tabcolsep}{10.54pt}
\renewcommand{\arraystretch}{1.0}
\begin{tabular}{@{}lcccccc@{}}
\toprule
 & \multicolumn{4}{c}{\textbf{Generation}} & \multicolumn{2}{c}{\textbf{IS}} \\
\cmidrule(lr){2-5}\cmidrule(lr){6-7}
Method & Contribution & Importance & Soundness & Novelty & Lit.\ novelty & Quality \\
 & (1--5) & (1--5) & (1--5) & (0--5) & (0--100) & (0--100) \\
\midrule
\multicolumn{7}{@{}l}{\textbf{GPT-6 Astra}}\\*
\hspace{0.4em}Raw-paper RAG & 2.99\,/\,2.58 & 3.27\,/\,2.94 & 3.38\,/\,3.16 & 3.48\,/\,2.81 & 48.60\,/\,46.70 & 72.50\,/\,60.20 \\
\rowcolor{brandbg}\hspace{0.4em}Single (ours) & 3.35\,/\,{\rmfamily\bfseries 3.02} & 4.12\,/\,3.24 & 3.58\,/\,3.47 & 3.92\,/\,3.48 & 57.74\,/\,55.26 & 81.40\,/\,75.00 \\
\rowcolor{brandbg}\hspace{0.4em}Multi (ours) & {\rmfamily\bfseries 3.41}\,/\,{\rmfamily\bfseries 3.02} & {\rmfamily\bfseries 4.18}\,/\,{\rmfamily\bfseries 3.26} & {\rmfamily\bfseries 3.93}\,/\,{\rmfamily\bfseries 3.65} & {\rmfamily\bfseries 4.00}\,/\,{\rmfamily\bfseries 3.55} & {\rmfamily\bfseries 59.89}\,/\,{\rmfamily\bfseries 57.37} & {\rmfamily\bfseries 82.10}\,/\,{\rmfamily\bfseries 76.70} \\
\midrule
\multicolumn{7}{@{}l}{\textbf{Qwen3.8-27B}}\\*
\hspace{0.4em}Raw-paper RAG & 2.69\,/\,2.18 & 2.71\,/\,2.31 & {\rmfamily\bfseries 3.08}\,/\,{\rmfamily\bfseries 3.05} & 2.42\,/\,2.10 & 38.90\,/\,38.50 & 70.40\,/\,62.50 \\
\rowcolor{brandbg}\hspace{0.4em}Single (ours) & 2.95\,/\,{\rmfamily\bfseries 2.76} & 3.32\,/\,2.67 & 2.71\,/\,2.79 & 2.83\,/\,{\rmfamily\bfseries 2.77} & 43.00\,/\,46.10 & 70.20\,/\,68.90 \\
\rowcolor{brandbg}\hspace{0.4em}Multi (ours) & {\rmfamily\bfseries 3.00}\,/\,2.75 & {\rmfamily\bfseries 3.37}\,/\,{\rmfamily\bfseries 2.73} & 2.74\,/\,2.80 & {\rmfamily\bfseries 2.95}\,/\,2.76 & {\rmfamily\bfseries 44.30}\,/\,{\rmfamily\bfseries 48.60} & {\rmfamily\bfseries 70.90}\,/\,{\rmfamily\bfseries 69.50} \\
\midrule
\multicolumn{7}{@{}l}{\textbf{GLM-5.3-Flash}}\\*
\hspace{0.4em}Raw-paper RAG & 2.80\,/\,2.50 & 3.04\,/\,2.66 & 3.26\,/\,3.37 & 3.04\,/\,2.72 & 42.80\,/\,44.60 & 71.40\,/\,67.50 \\
\rowcolor{brandbg}\hspace{0.4em}Single (ours) & 3.07\,/\,2.94 & 3.83\,/\,3.18 & 3.27\,/\,3.32 & 3.82\,/\,3.31 & 48.00\,/\,53.40 & 71.00\,/\,71.50 \\
\rowcolor{brandbg}\hspace{0.4em}Multi (ours) & {\rmfamily\bfseries 3.13}\,/\,{\rmfamily\bfseries 3.00} & {\rmfamily\bfseries 3.91}\,/\,{\rmfamily\bfseries 3.23} & {\rmfamily\bfseries 3.50}\,/\,{\rmfamily\bfseries 3.40} & {\rmfamily\bfseries 3.90}\,/\,{\rmfamily\bfseries 3.37} & {\rmfamily\bfseries 48.60}\,/\,{\rmfamily\bfseries 53.50} & {\rmfamily\bfseries 71.60}\,/\,{\rmfamily\bfseries 72.20} \\
\midrule
\multicolumn{7}{@{}l}{\textbf{Qwen3.8-Flash-Next}}\\*
\hspace{0.4em}Raw-paper RAG & 2.72\,/\,2.36 & 2.83\,/\,2.45 & {\rmfamily\bfseries 3.44}\,/\,{\rmfamily\bfseries 3.28} & 2.63\,/\,2.18 & 42.60\,/\,44.40 & 70.10\,/\,62.40 \\
\rowcolor{brandbg}\hspace{0.4em}Single (ours) & 2.96\,/\,2.78 & 3.44\,/\,2.75 & 3.26\,/\,3.12 & 2.96\,/\,2.95 & 47.80\,/\,50.60 & {\rmfamily\bfseries 72.90}\,/\,69.60 \\
\rowcolor{brandbg}\hspace{0.4em}Multi (ours) & {\rmfamily\bfseries 3.00}\,/\,{\rmfamily\bfseries 2.90} & {\rmfamily\bfseries 3.50}\,/\,{\rmfamily\bfseries 2.82} & 3.30\,/\,3.09 & {\rmfamily\bfseries 3.04}\,/\,{\rmfamily\bfseries 3.10} & {\rmfamily\bfseries 48.30}\,/\,{\rmfamily\bfseries 51.30} & 71.70\,/\,{\rmfamily\bfseries 70.70} \\
\midrule
\multicolumn{7}{@{}l}{\textbf{Average}}\\*
\hspace{0.4em}Raw-paper RAG & 2.80\,/\,2.41 & 2.96\,/\,2.59 & 3.29\,/\,3.22 & 2.89\,/\,2.45 & 43.23\,/\,43.55 & 71.10\,/\,63.15 \\
\rowcolor{brandbg}\hspace{0.4em}Single (ours) & 3.08\,/\,2.88 & 3.68\,/\,2.96 & 3.21\,/\,3.18 & 3.38\,/\,3.13 & 49.14\,/\,51.34 & 73.88\,/\,71.25 \\
\rowcolor{brandbg}\hspace{0.4em}Multi (ours) & {\rmfamily\bfseries 3.14}\,/\,{\rmfamily\bfseries 2.92} & {\rmfamily\bfseries 3.74}\,/\,{\rmfamily\bfseries 3.01} & {\rmfamily\bfseries 3.37}\,/\,{\rmfamily\bfseries 3.24} & {\rmfamily\bfseries 3.47}\,/\,{\rmfamily\bfseries 3.20} & {\rmfamily\bfseries 50.27}\,/\,{\rmfamily\bfseries 52.69} & {\rmfamily\bfseries 74.08}\,/\,{\rmfamily\bfseries 72.28} \\
\bottomrule
\end{tabular}
\floatgap
\end{table}

\looseness=-1 For ablation, we construct a raw-paper retrieval baseline (RAG; Table~\ref{tab:ablation}), which, like a web search, retrieves papers from the same sources as InsightPro and generates one idea from the retrieved text, but cannot see the insights and problems that InsightPro synthesizes from these papers.

\looseness=-1 With Astra as the generator, Single and Multi exceed RAG in all six dimensions under both judges.
Averaged over the four generation models with RAG results, Multi is higher in Contribution, Importance, Soundness, and Novelty under both judges, and Single is higher in Contribution, Importance, and Novelty; in Importance, both workflows are higher than RAG with every generation model.
Literature novelty is higher for both workflows with every generation model and under both judges (on average 51.5 for Multi against 43.4 for RAG, averaged over the judges), and quality is also higher on average for both workflows.

Soundness varies with the generation model: RAG is higher with both Qwen3.8 models, while Multi is higher with Astra and GLM-5.3-Flash; the paired Card--Blueprint comparison above shows that this dimension mainly reflects how detailed the proposal description is.

\secgap\subsection{Case Studies: Research Ideation in Model Efficiency}
\label{sec:specific}

\looseness=-1 We examine how the six research perspectives in Section~\ref{sec:zh-ideax} organize research directions within model efficiency.
The cases span low-precision training, model compression, speculative decoding, and MoE memory management, showing how literature connections become concrete designs: randomized estimation motivates matched-state replay for compressed backpropagation; weight--scale coupling motivates controlled FP4 replay; and the distinction between verification demand and future reuse motivates rollback-aware expert residency.
The complete Idea Cards of these cases are shown on our demo website (\url{https://insightpro-demo.einsia.ai}).

\secgap\section{Conclusion}
\label{sec:conclusion}

The central idea behind InsightPro is to draw on how human researchers accumulate and recall knowledge: reading and comparing papers to distill reusable insights from scattered findings and identify worthwhile problems from the limitations of existing methods.
We implement this approach in a literature database organized by domain, compressing repeated content while retaining mechanisms, conditions of validity, and sources, and supplying this knowledge to agents' contexts.
Prompts built on the database guide agents to connect problems with insights, develop candidate mechanisms, and check prior work, delivering Idea Cards and paired Research Blueprints for researchers to examine and refine.

\looseness=-1 We believe research ideation rests on two foundations: the model's reasoning ability and the quality of the knowledge in its context.
Knowledge quality includes relevance to the target problem, the density of useful knowledge, accuracy of statements and their supporting evidence, and clear organization of mechanisms and applicability conditions.
Compared with accumulating raw papers retrieved by an agent, synthesizing insights and problems across papers aims to reduce repeated background and irrelevant detail, concentrating limited context on important obstacles, useful findings, and their evidence to support grounded problem selection and mechanism design.

Across 38 domains, single-agent and multi-agent generation workflows based on the database lead nine adapted baselines in contribution, importance, novelty, and literature novelty under two independent judges and are in the first tier in soundness and proposal quality.
Across generation models, scores follow a consistent ordering (GPT-6 Astra $>$ GLM-5.3-Flash $\approx$ DeepSeek-V4.1-Flash-1017 $\geq$ Qwen3.8-Flash-Next $>$ Qwen3.8-27B), and our workflows lead in contribution and importance under each of them, in line with the view above that ideation depends on both the model and the knowledge in its context.
Blueprint elaboration raises soundness, literature novelty, and proposal quality while leaving contribution and importance unchanged, and historical reconstruction provides a complementary test.
These results support the value of the database and its use.
Improving model capabilities and improving the research knowledge supplied to models are two directions that we believe should advance together.

\FloatBarrier
\secgap\section{Related Work}
\label{sec:related}

Research ideation with language models begins with finding useful evidence and inspiration in prior work. Retrieval-augmented generation brings external information into a model's context~\citep{lewis2020rag}, while OpenScholar retrieves scientific passages and synthesizes citation-backed answers to help researchers access domain knowledge~\citep{openscholar2024}. For ideation, SciMON retrieves inspirations from earlier studies and iteratively revises candidates through comparisons with prior work~\citep{scimon2023}; MOOSE combines research backgrounds, inspirational material, and multiple forms of feedback to develop hypotheses~\citep{moose2023}. SciPIP builds a paper database containing keywords, backgrounds, ideas, concise methods, and references, combines multiple retrieval strategies, and integrates proposals derived from literature and the model's internal knowledge~\citep{scipip2024}. These approaches progressively connect literature access with the development of research ideas.

To make borrowing from the literature more concrete, several methods decompose papers into components that can be compared and combined. Scideator extracts purposes, mechanisms, and evaluation facets, supporting their analogical recombination by researchers alongside novelty checks~\citep{scideator2024}. IdeaSynth represents problems, solutions, evaluations, and contributions as editable idea facets, using literature-grounded feedback for iterative refinement and composition~\citep{ideasynth2024}. Work on combinatorial creativity retrieves across abstraction levels to discover connections between domains, then analyzes and recombines components of prior work~\citep{combinatorial2024}. SCI-IDEA similarly extracts key paper content and combines prompting strategies with token and sentence representations to support iterative ideation~\citep{sciidea2025}. These studies show that transferable knowledge units and the ways they are adapted are already an important entry point for research ideation.

Relationships among these units can also support new ideas. ResearchAgent augments a core paper with publications connected through an academic graph and concepts shared across papers, then uses reviewing agents to revise problems, methods, and experimental designs~\citep{researchagent2024}. CoI organizes literature into chains of research development~\citep{li2024coi}, while Tree-of-Ideas reconstructs branching trajectories and reasons across them to identify shared problems and complementary solutions~\citep{treeideas2026}. SciAgents combines conceptual connections in a knowledge graph with multi-agent discussion to generate and refine hypotheses~\citep{sciagents2024}. Graph2Idea retrieves papers for an input topic, extracts knowledge triples, and dynamically constructs a graph that supplies compact relational context for candidate generation~\citep{graph2idea2026}. Structured knowledge, scholarly evolution, and connections across fields thus inform ideation in several forms.

\looseness=-1 Once inspirations are available, a system must decide how to combine them, which candidates to retain, and how far to develop each proposal. MOOSE-Chem decomposes hypothesis discovery into inspiration retrieval, hypothesis composition, and ranking~\citep{moosechem2024}; MOOSE-Chem2 starts from a coarse research direction and uses hierarchical search to progressively add key methodological and experimental details~\citep{moosechem22025}. Nova iteratively plans external knowledge retrieval to broaden the novelty and diversity of candidates~\citep{nova2024}. IRIS combines Monte Carlo Tree Search, literature synthesis, and researcher feedback, allowing users to steer idea development~\citep{iris2025}. These methods treat knowledge acquisition and proposal elaboration as decisions that can be revisited.

\looseness=-1 Search over multiple candidates often combines collaboration with selection. VirSci coordinates roles that jointly generate, evaluate, and refine ideas~\citep{virsci2024}, while AI co-scientist uses generation, debate, and tournament-based evolution to improve hypotheses under scientist-provided objectives~\citep{aicoscientist2025}. FlowPIE couples literature exploration with candidate evolution~\citep{flowpie2026}, and EvoGens combines retrieval planning, mutation, and conceptual recombination to sustain exploration within a population of ideas~\citep{evogens2026}. IDEAgent jointly considers quality and diversity, using candidate lineages, repair feedback, and comparisons with past ideas to guide further generation~\citep{ideagent2026}. These designs provide different ways to implement discussion and selection, and they also mean that comparing systems requires jointly accounting for candidate counts, knowledge access, and computational budgets.

Training offers another route to improving how models use knowledge and judge candidates. Sparks of Science organizes conventional assumptions, key insights, counterproposals, and their reasoning connections from papers to train hypothesis generation~\citep{sparksofscience2025}; Graphs of Research uses citation-evolution relationships as supervision for idea generation~\citep{graphsofresearch2026}. MOOSE-Star decomposes and trains inspiration retrieval and hypothesis composition~\citep{moosestar2026}, while Rubric Rewards uses research objectives and grading criteria to train research-plan generation~\citep{rubricrewards2025}. Spark incorporates an evaluator trained on scientific reviews~\citep{spark2025}, and CycleResearcher couples research and reviewing models, using feedback-based training to improve research outputs~\citep{cycleresearcher2024}. IdeaScientist organizes gap finding, mechanism transfer across problems, and report writing into trained roles and provides structured literature records with research results masked~\citep{liu2026ideascientist}. These works suggest that model training and knowledge supply at generation time are two directions of improvement that can be combined.

When data and experiments are available, ideation can also receive feedback from the phenomena being studied. HypoGeniC generates linguistic hypotheses from examples and updates candidates using predictive performance~\citep{hypogenic2024}; HypoRefine combines findings from literature with patterns in data to refine hypotheses~\citep{hyporefine2024}. The AI Scientist connects ideation to implementation, experiments, manuscript writing, and review~\citep{aiscientist2024}, and Dolphin feeds experimental analyses into subsequent ideation~\citep{dolphin2025}. Kosmos uses a structured world model to accumulate information across literature search, data analysis, and hypothesis generation, linking report claims to code or literature evidence~\citep{kosmos2025}. HypER focuses on literature-grounded hypotheses and their reasoning chains, training models to distinguish valid from invalid scientific reasoning~\citep{hyper2025}. These studies connect proposal formation, evidence records, and subsequent validation.

Evaluating ideation consequently requires distinguishing proposal quality, knowledge use, and research outcomes. Si et al. compare model and human ideas through blinded expert review~\citep{si2024ideation}, then follow ideas through implementation in a subsequent study to compare assessments before and after execution~\citep{si2025execution}. ResearchBench separately evaluates inspiration retrieval, hypothesis composition, and ranking~\citep{researchbench2025}; AI Idea Bench combines comparisons with published research content and judgments based on reference material~\citep{aiideabench2025}. FIRE-Bench further requires agents to conduct experiments from a research question and produce empirically supported conclusions to assess scientific rediscovery~\citep{firebench2026}. Such evaluations of complete research processes provide a different level of evidence from proposal assessment before experimentation.

\looseness=-1 The reliability of evaluation models requires independent examination. NovBench tests whether models produce relevant and accurate novelty assessments relative to expert reviews~\citep{novbench2026}; RQ-Bench compares model and expert judgments of research questions and finds their preferences can diverge~\citep{rqbench2026}. SoundnessBench examines methodological soundness assessable at the proposal stage and models' tendency toward excessive optimism~\citep{soundnessbench2026}, while SciStyleBench controls scientific content and changes presentation to measure score shifts~\citep{scistyle2026}. These findings motivate reporting the two judges separately, comparing them with human experts, and analyzing how scores change as the same idea is elaborated from an Idea Card into a Research Blueprint.

\looseness=-1 Building on this literature, InsightPro focuses on condensing domain literature into reusable research knowledge before ideation. During offline construction, we aggregate insights and problems across papers while retaining mechanisms, applicability conditions, and sources. Agents then retrieve this knowledge through an interface, so that it provides context for ideation. The emphasis is on improving the representation and coverage of the knowledge available to agents when they select problems and establish mechanistic connections; this resource can be combined with different generation models, candidate search procedures, and collaboration workflows. Idea Cards and their accompanying Research Blueprints retain the core idea, its rationale, and the elaborated design for subsequent examination and validation.

\FloatBarrier
\secgap\section*{Limitations and Future Work}
\label{sec:limitations}

The results show that the full \method{} workflow is useful in practice and that elaborating proposals changes the apparent strengths of a system.
Because evaluating research ideas is inherently difficult, we do not implement the proposed ideas and measure them on downstream benchmarks; like prior work~\citep{si2024ideation,li2024coi,liu2026ideascientist}, we evaluate them with rubric-based LLM judges, whose scores reflect proposal quality under these rubrics rather than verified scientific contribution.
The results also do not separate whether the gains come from selecting better problems, drawing on additional mechanisms, organizing knowledge, or the generation and selection budget.
Prompt development involved the efficiency, latent-reasoning, and distillation domains, which are also among the 38 evaluation domains.
Future work includes executing selected ideas and measuring them on downstream benchmarks, building better multi-agent systems to further improve performance, and controlled comparisons that isolate the knowledge representation, for example a same-source summary baseline with a comparable number of tokens, and the amount of context.

\par\label{m2p:end-main}\typeout{M2P-ENDMAIN remaining=\the\dimexpr\pagegoal-\pagetotal\relax\space baselineskip=\the\baselineskip}
\newpage
\bibliographystyle{unsrtnat}
\begingroup
\renewcommand{\bibfont}{\fontsize{9}{11}\selectfont}
\setlength{\bibsep}{2pt plus .5pt}
\bibliography{ref}  
\endgroup
\clearpage
\appendix
\section{Prompts and Output Specifications}
\label{app:prompts}

The runtime generation and writing prompts used in the reported experiments are in Chinese.
Section~\ref{app:single-prompt} provides a complete English single-agent prompt for reuse, together with its required inputs.
The remaining subsections give condensed English descriptions of coordination, writing, and evaluation instructions.
The original Chinese evaluation prompts are available in the code repository.
Dynamic domain names, dates, and the service address are instantiated at run time.

\subsection{Single-Agent Generation: Complete Reusable Prompt}
\label{app:single-prompt}

The prompt below is the single-agent prompt used in the experiments (Section~\ref{sec:blueprint}), translated from Chinese.
It retains the research instructions and output contract, while replacing deployment-specific client commands with documented HTTP operations and configurable paths.
In the experiments, five isolated sessions per domain each run the prompt below and deliver only their top card (Table~\ref{tab:system-config}).
Public web access is optional in this version; the reported experiments ran without it.

Table~\ref{tab:prompt-inputs} lists the placeholders in the prompt; the released dataset includes a launcher for running the InsightPro service locally~\citep{einsia2026insightpro}.

\begin{longtable}{@{}p{.37\linewidth}p{.59\linewidth}@{}}
\caption{Inputs for the reusable single-agent prompt. Replace every occurrence of each placeholder.}\label{tab:prompt-inputs}\\
\toprule
Placeholder & Value to supply \\
\midrule
\endfirsthead
\toprule
Placeholder & Value to supply \\
\midrule
\endhead
\bottomrule
\endfoot
\path{{current_date}} & The date of the generation run. \\
\path{{domain_name}}, \path{{domain_id}} & The assigned domain's name and actual InsightPro identifier. \\
\path{{api_base}} & The accessible service root URL, without a trailing slash or \path{/api/v1} suffix. \\
\path{{output_dir}} & An absolute writable directory, such as \path{/work/out}. \\
\end{longtable}

\clearpage
\begin{tcblisting}{listing only,breakable,colback=white,colframe=brandprimary,
  title={Single-agent generation prompt (English translation)},
  fonttitle=\bfseries\sffamily,coltitle=brandfg,colbacktitle=brandbg,
  boxrule=.4pt,arc=2pt,left=6pt,right=6pt,top=5pt,bottom=5pt,
  before skip=8pt,after skip=8pt,
  listing options={basicstyle=\ttfamily\small,columns=fullflexible,
    keepspaces=true,breaklines=true,breakatwhitespace=true,breakindent=0pt,
    showstringspaces=false,showspaces=false,showtabs=false,tabsize=2,
    upquote=true,numbers=none}}
BEGIN INSIGHTPRO SINGLE-AGENT PROMPT

1. ROLE, TASK, AND INPUTS

You are an independent researcher. This is a single-agent research idea generation task. Deliver the one new research idea most worth developing into an independent paper. Do not call sub-agents. You may develop, compare, and discard candidates internally, but deliver only one complete idea.

Current date: {current_date}
Domain: {domain_name}
Domain ID: {domain_id}
InsightPro service root: {api_base}
Writable output directory: {output_dir}

Within the assigned domain, choose important research problems and technical approaches yourself. Develop candidate ideas and deliver the strongest one. There is no additional historical cutoff for literature retrieval.

InsightPro organizes AI papers from the last ten years by research domain. From these papers, it summarizes recurring insights and problems. Insights include mechanisms, regularities, reusable understanding, and their applicability conditions. Problems include important recurring questions, bottlenecks, and unresolved challenges. These entries are starting points for exploration and link back to source papers.

2. INVESTIGATE THE DOMAIN

Use the InsightPro API actively to investigate problems, mechanisms, and source papers as your research requires; you decide the scope, order, and depth of retrieval. You may read a complete domain when needed, but reading every entry is not a required first step.

The library summarizes existing literature. A historical problem may already have been solved; an existing insight is not itself your new contribution. Check relevant work before making either judgment.

Each idea must aim to address at least one core problem in the assigned domain. Explain why it matters, the key obstacle in existing approaches, and how your proposed mechanism acts on that obstacle. Give a grounded proposal; you are not required to have experimentally proved it at the idea-generation stage. You may start from a problem and seek a useful insight, or start from an insight and discover a way to address a core problem.

Use insights from this domain or any other domain. Cross-domain connections and repeated exploration are encouraged. When another domain's insights or problems become important to your proposal, read the relevant entries, source papers, or that domain's entries as needed to understand its mechanisms, assumptions, and problem context. Choose cross-domain connections yourself; there is no prescribed shortlist or coverage quota.

When an insight's meaning, applicability, or evidence needs checking, retrieve its linked papers or search for related work by mechanism and problem. Follow source papers to nearby work when useful. Distinguish what you actually read from your own deductions. A paper ID listed in an entry does not mean that you have read that paper's full text. Explain which problems and insights motivated each final idea and how they informed its design.

3. SCIENTIFIC AMBITION

Bold but grounded attempts are encouraged. Develop ideas that, if successful, could enable important capabilities, methods, or understanding. Do not choose only safe, easy-to-implement, or easy-to-write incremental changes. Grounds for exploration may come from real literature, observed phenomena, reasonable deductions, or structural connections across domains. Such grounds support exploration; they do not prove a new hypothesis.

Do not judge an idea negatively or discard it merely because it has not been tried, has no completed experiments, has an incomplete experimental plan, or has unresolved data and resource arrangements. For a bold proposal, explain the concrete mechanism, why it is worth trying, its key assumptions, unresolved steps, and testable predictions. Allow reasonable uncertainty and first try developing the mechanism, an alternative construction, or a more valuable problem formulation.

Continue to scrutinize existing directions. Confirmed core duplication, factual errors, and logical contradictions require substantive correction; ambition does not excuse them.

4. SIX RESEARCH PERSPECTIVES

Use the following perspectives to explore the literature library, find insights, and develop ideas. They can be combined, revisited, or extended. They are not a checklist to complete in order. Focus on finding a new way to address a core problem, not on inventing an example for a rule.

G01 - A+B and complementary mechanisms. Start from an important problem and find mechanisms that can jointly remove a bottleneck. Explain why A and B are complementary, how they interact, and what independent value their combination can add. Familiar components can support a new contribution, but "nobody has combined A and B" is not sufficient evidence of novelty.

G02 - Cross-domain transfer. Find structural connections between the current problem and a method, mechanism, or insight from another domain. Explain why it works in the source setting, which assumptions still hold here, and what adaptation is required. Judge the transfer by the new capability or understanding it enables, not by how many components were changed.

G03 - Rethinking default assumptions. Reexamine accepted assumptions, problem definitions, or explanations. Investigate their evidence and applicability boundaries. Consider what new methods, predictions, or understanding could follow from relaxing, replacing, or correcting them. The conclusion need not overturn prior work.

G04 - Diagnosis, remeasurement, and reattribution. Start from anomalies, conflicting evidence, or unexplained regularities. Identify the factors that actually matter, propose measurements or controls that distinguish competing explanations, and consider improvements suggested by the diagnosis. Positive and negative findings can both be valuable when they answer an important question and add understanding.

G05 - Extending capabilities and applicability. Examine whether a method's operating unit, adaptive variables, representation, or theoretical scope can be extended. Explain what causes the existing boundary, how it could be crossed, and what capability might result. Choose constructions appropriate to the problem rather than restricting yourself to one parameter change or the smallest possible modification.

G06 - Free exploration. Start directly from an important unresolved problem or observation in the domain. Develop constructions beyond the preceding perspectives or combine several of them. You do not need to prove that the other rules are inapplicable. Still explain the actual basis, substantive contribution, and testable predictions.

The final idea does not need to use all six perspectives. Select among candidates by problem value, potential contribution, concrete mechanism, evidence, and novelty. Do not impose quotas on idea type or risk level.

5. LITERATURE SEARCH AND READING

Public web access is optional; decide whether it is needed. Actually retrieve and read nearby work through InsightPro and/or public literature before finalizing ideas; do not claim novelty from memory alone. Public paper databases, arXiv, OpenReview, publisher pages, and ordinary web pages may be used. Use terminal HTTPS requests, such as curl or Python, to access search services and sources. Crossref and DataCite are possible public metadata search services. Change keywords or sources when retrieval is unhelpful.

Use at least three distinct query formulations to investigate related mechanisms, existing combinations, and the strongest relevant baselines. Actually read the most relevant sources. An empty search result is not proof of novelty. Actually use the InsightPro API during this run; public web access is optional, and not using it does not make the run incomplete.

Do not access unrelated local files, earlier outputs, or other sessions. The explicitly supplied InsightPro service is an authorized resource for this task.

Investigate the core mechanisms and conclusions that previous work has already established. State your substantive remaining contribution. Combining familiar components or transferring a mechanism across domains can yield a contribution, but renaming, repeating known conclusions, or relying only on the absence of a previously tried combination does not establish novelty.

For the final idea, state the problem, core mechanism and any necessary formulas, a causal or falsifiable prediction, the difference from the closest work, and why the contribution could support an independent paper.

The experiment section may be a short proposal for distinguishing the core hypothesis from existing explanations: what could be observed, and what results would make you revise your judgment. Suggest datasets, tasks, metrics, controls, and resources only as needed, and state what remains undecided. Do not require a complete experimental protocol, budget, or existing results at this stage. Do not use experimental completeness as a candidate-admission or ranking criterion.

Do not fabricate experimental results. Distinguish literature facts, your deductions, unverified hypotheses, and risks that retrieval has not ruled out.

6. NOVELTY CHECK AND HONEST SELF-ASSESSMENT

After developing candidates and before choosing the final idea, check each candidate's novelty and honestly assess whether its contribution could grow into an independent research paper.

First state the actual core contribution in one or two sentences: which problem it addresses, which mechanism, capability, or understanding it proposes, and what it adds beyond existing work. Search InsightPro and public literature for the closest predecessors using that core contribution, different terminology, and equivalent formulations. Actually read key sources. Compare the research question, core operations, mechanism, and main conclusions. Different titles, names, or application settings alone do not establish a difference.

If prior work already achieves the same core contribution, explicitly acknowledge the overlap, substantively revise or replace the candidate, and check it again. Do not continue claiming that contribution after merely renaming it, rephrasing it, or adding secondary details. If an independent contribution remains, identify it precisely, explain why it matters, and assess whether it is sufficient for a paper. Record unresolved duplication risks honestly; do not downplay the closest predecessor to preserve your candidate.

Ask yourself: if the core proposal succeeds, what important capability or understanding would it add? Is the remaining contribution substantive and independent of the strongest relevant work? What is its strongest basis, its weakest link, and the critical hypothesis that remains unverified?

Do not give every candidate an automatically positive assessment or promise acceptance at a venue. If the contribution is insufficient, develop the mechanism further or replace the candidate. If it has paper potential but important unknowns remain, give a conditional judgment. Assess scientific potential separately from experimental completeness; lack of results, unsettled experimental details, or an untried idea is not by itself a reason to reject its potential.

Record actual searches, closest prior work, overlaps, remaining contributions, and unresolved risks in novelty_check. In scientific_potential, give an honest conclusion, concrete reasons, known limitations, and the key conditions required for success. Present these judgments in IDEA.md as well. Support self-assessment with comparisons and reasoning rather than substituting a self-assigned score for evidence.

7. REQUIRED DELIVERABLES

Write the following files inside {output_dir}:

1) IDEAS.json: a valid JSON array containing exactly one object. The object must have these nonempty fields:
title, problem, mechanism, previous_work, novelty_check, experiment, failure_conditions, scientific_potential.

previous_work must contain at least three of the most relevant real works. Each record must include:
title (complete title), url (actual source), read_depth, source_location (the abstract, section, or field actually read), overlap, difference.

novelty_check must contain:
search_queries, closest_work, already_done_test, remaining_contribution, unresolved_risks.

problem must identify at least one core domain problem by its real InsightPro entry ID, explain its importance, and state the unresolved obstacle. mechanism must identify the actual insight entries used, their source domains, and how they inspired the proposed solution. Never invent library IDs.

experiment may be a brief future validation plan with undecided elements stated explicitly. failure_conditions must describe observations or conditions that could refute the core hypothesis; completed experiments are not required.

2) IDEA.md: a readable Chinese explanation for other researchers. Fully explain the idea, its previous work, and proposed validation, and explain why it was chosen among the candidates you explored.

3) SEARCH_LOG.md: actual queries, source URLs, reading depth, candidate choices and rejections, and unresolved risks. Record tool failures honestly.

You may store retrieved material and scripts inside {output_dir}; treat the rest of the supplied workspace as read-only. Before finishing, validate the JSON and check that all three deliverables are complete.

The delivered idea must satisfy these requirements. Do not bundle several independent proposals into one idea.

8. READ-ONLY INSIGHTPRO ACCESS

The live API provides entries and original evidence on demand. Actually use it to investigate the assigned domain, nearby entries, and source papers. paper_count means the number of supporting papers, not the number of external citations.

The service root is {api_base}. Read its /llms.txt or /openapi.json documentation when needed. Use HTTPS requests to the following routes; URL-encode query values and replace example IDs with real IDs:

GET /api/v1/stats
GET /api/v1/domains?q=QUERY&limit=20
GET /api/v1/domains/DOMAIN_ID
GET /api/v1/domains/DOMAIN_ID/insights?limit=100
GET /api/v1/domains/DOMAIN_ID/problems?limit=100
GET /api/v1/domains/DOMAIN_ID/papers
GET /api/v1/search?q=QUERY&kind=insight&mode=all&limit=20
GET /api/v1/search?q=QUERY&kind=problem&mode=any&limit=20
GET /api/v1/entries/ENTRY_ID
GET /api/v1/papers/NUMERIC_PAPER_ID
POST /api/v1/papers/batch
  JSON body: {"ids": [NUMERIC_PAPER_ID_1, NUMERIC_PAPER_ID_2]}

The batch POST performs read-only lookup. These routes can be called directly; no experiment-specific client script is required. For example, replace SEARCH_TERMS with a mechanism or problem description:

curl -G "{api_base}/api/v1/search" \
  --data-urlencode "q=SEARCH_TERMS" \
  --data-urlencode "kind=insight" \
  --data-urlencode "mode=all" \
  --data-urlencode "limit=20"

To read all entries of a domain, first read its domain record and counts. Fetch every page of both insights and problems. A successful response has ok=true and a data field. For paginated entry lists, inspect page.has_more; if true, pass page.next_after as the next request's after parameter. Continue until has_more=false. Reject empty or repeated cursors, duplicate IDs, and incomplete pages. Deduplicate by entry ID and verify that the final insight and problem counts match the domain record. Recheck the counts after retrieval to detect changes during retrieval.

Save the retrieved entries as JSONL inside {output_dir}, then actually read every entry, in chunks if necessary. Download completion is not evidence of reading. Report missing materials or incomplete reading honestly.

For previous_work records verified through the API, also provide paper_id. In IDEA.md, present the numeric paper_id together with the complete title so that the source can be checked. An InsightPro numeric paper ID is not an OpenAlex W identifier. You may cite papers outside the library, but must not invent an InsightPro ID for them.

Record actual API queries and reading depth in SEARCH_LOG.md. If the API is unavailable, report the failure; do not silently fall back to a local database or pretend that the requested evidence was read. Entry IDs must come from evidence actually returned by retrieval in this run; do not invent them.

END INSIGHTPRO SINGLE-AGENT PROMPT
\end{tcblisting}

\subsection{Multi-Agent Coordination}

Each of six authors uses the shared generation prompt and produces five cards.
The candidate pool is frozen before one discussion--response cycle.
Authors may respond to criticism, retain a bold but supported direction, revise their own proposal, or withdraw it.
They may not silently rewrite another author's card.
The selector receives the resulting candidates and chooses five existing identifiers in order.

\subsection{Research Blueprint}

\begin{promptbox}{Writing-stage principles}
Expand the selected Idea Card while preserving its core problem, scientific claim, mechanism, contribution, and scope. Explain both why the idea was proposed and how it would work.

Reuse prior research records and evidence. Make a targeted new query only when a specific conflict, missing factual basis, or new claim needs checking. Do not pretend that a later reconstruction was part of the original research record.

Distinguish literature facts, deductions, proposed design choices, and unverified hypotheses. Label new details. If a central contradiction or prior-work overlap requires changing the mechanism, flag a return to ideation instead of replacing it silently.
\end{promptbox}

The blueprint contains the title and core claim; the chain from problem and prior-work obstacle to insight, adaptation, and design; a problem definition; a detailed method; a concrete example; comparison with the closest work; an initial validation plan; contribution and boundaries; and a statement of what was added relative to the card.
Formulas or pseudocode are used where they remove central ambiguity, not as compulsory decoration.
The plan identifies what observations would support, contradict, or leave the claim unresolved.
The writer checks that the mechanism is preserved.
Unspecified budgets or final hyperparameters remain open rather than being invented.

\subsection{Generation Judge Instructions}

For Contribution, Importance, and Soundness, the judge asks three questions: what remains new relative to close work; why achieving the objective matters; and whether the stated premises, mechanism, and expected outcome are coherent.
The judge uses available evidence, keeps uncertainties distinct from demonstrated flaws, and does not repair the idea before scoring it.

For conceptual Novelty, the judge scores each card on its own.
It notes the mechanism, strength, limitation, and a supporting quotation, then scores the conceptual departure.
The protocol does not force a ranking, quota, or score distribution and does not treat the author's own novelty claim as independent evidence.

\section{Additional Experimental Comparisons}
\label{app:experiments}

Table~\ref{tab:system-config} lists the source, native workflow, literature access, and adaptations of every system compared in the main text.

\begingroup
\footnotesize
\setlength{\tabcolsep}{5.04pt}
\renewcommand{\arraystretch}{1.15}
\begin{longtable}{@{}>{\raggedright\arraybackslash}p{0.15\linewidth}>{\raggedright\arraybackslash}p{0.2\linewidth}>{\raggedright\arraybackslash}p{0.28\linewidth}>{\raggedright\arraybackslash}p{0.30\linewidth}@{}}
\caption{\textbf{Configurations of the compared systems.} Every system receives the same domain instruction and delivers five independent final outputs per domain; native candidate generation, critique, and selection stay inside each system, and every output starts from a fresh native context. Unless noted otherwise, baselines run offline: public network access is blocked, the generation model is reached through an API bridge, and literature comes from one frozen common corpus of 283,008 raw paper records (282,988 with abstracts and 1,948 with local full text) through local search and read tools. Links point to the frozen upstream commits; for our workflows, to the released code.}\label{tab:system-config}\\
\toprule
System & Source (GitHub) & Native workflow and final output & Literature access and adaptations \\
\midrule
\endfirsthead
\multicolumn{4}{@{}l}{\textit{Table continued}}\\
\toprule
System & Source (GitHub) & Native workflow and final output & Literature access and adaptations \\
\midrule
\endhead
\midrule\multicolumn{4}{r@{}}{\textit{Continued on next page}}\\
\endfoot
\bottomrule
\endlastfoot
IdeaScientist & Astra: \href{https://github.com/jiarui-liu/IdeaScientist/tree/82a98a08772eab08d11fd097445afa4a8eaaa5b3}{\nolinkurl{github.com/jiarui-liu/IdeaScientist}} (\texttt{82a98a0}); others: \href{https://github.com/Einsia/InsightPro/tree/71d3b5f7ace981fb43876a497dae708116edde87/baselines/ideascientist}{\nolinkurl{github.com/Einsia/InsightPro}} (\texttt{71d3b5f}) & Four native roles (gap finder, innovator, reviewer, report writer) that keep their revise and reject decisions; one proposal per run. & Astra: its own paper library plus full text fetched from public arXiv, without a date cutoff. Other generators: the common corpus only; its own library is not used. \\
CoI & Astra: \href{https://github.com/DAMO-NLP-SG/CoI-Agent/tree/ac94317ff2d1997f362747edc2c9e5f0fdc9c450}{\nolinkurl{github.com/DAMO-NLP-SG/CoI-Agent}} (\texttt{ac94317}); others: \href{https://github.com/Einsia/InsightPro/tree/71d3b5f7ace981fb43876a497dae708116edde87/baselines/coi}{\nolinkurl{github.com/Einsia/InsightPro}} (\texttt{71d3b5f}) & One chain of two to three papers, then trend analysis, idea generation with novelty screening, experiment design, and at least one ReviewAgent revision; one proposal per run. & Online arXiv and OpenAlex search without a date cutoff for every generator (for the non-Astra generators, through a proxy that allows only public HTTP and HTTPS). Model-call timeout: 1,200\,s for Astra and 6,300\,s for the other generators. \\
AutoResearchClaw & \href{https://github.com/aiming-lab/AutoResearchClaw/tree/be4ba4755bf1b52220f25e13b2293b5956590070}{\nolinkurl{github.com/aiming-lab/AutoResearchClaw}} (\texttt{be4ba47}) & Native stages 1--9 up to experiment design: query expansion, paper screening, knowledge extraction, three hypothesis perspectives, and their synthesis. Final output: the hypothesis set with its experiment design. & Academic search, web search, and PDF retrieval all resolve to the common corpus; the seminal-paper list built into the code is kept and labelled as built-in prior knowledge. \\
Dr.~Claw & \href{https://github.com/OpenLAIR/dr-claw/tree/37493cd93b12fbf34bed68c6c6402b6e19aaf6d2}{\nolinkurl{github.com/OpenLAIR/dr-claw}} (\texttt{37493cd}) & Native idea-generation skill: three candidates, rereading in a fresh context, and selection against five gate criteria. & The 30-turn agent limit is removed; its web search and fetch tools query the common corpus. \\
DeepScientist & \href{https://github.com/ResearAI/DeepScientist/tree/b36624417f0c6b8238ec02db37b94d6db2faa5b0}{\nolinkurl{github.com/ResearAI/DeepScientist}} (\texttt{b366244}) & Native idea stage (quest creation, candidate drafts, comparison, and selection) with the baseline-reproduction step waived. Final output: the selected idea. & Native live web search is disabled; the agent uses the local search and read tools. \\
ResearchAgent & \href{https://github.com/JinheonBaek/ResearchAgent/tree/babb49b51ebfcebedc39ccde12c9785be7bef46c}{\nolinkurl{github.com/JinheonBaek/ResearchAgent}} (\texttt{babb49b}) & Native pipeline with three iterations of problem, method, and experiment refinement, each judged by validators on five criteria. Final output: the native proposal. & Seed papers, references, and entities come from the common corpus, encoded locally with SPECTER2. \\
SciAgents & \href{https://github.com/lamm-mit/SciAgentsDiscovery/tree/c5c30451b29cba813a11b5ce078909a214dad9f2}{\nolinkurl{github.com/lamm-mit/SciAgentsDiscovery}} (\texttt{c5c3045}) & Native group of 13 agents (at most 50 rounds, automatic speaker selection) that follows a path in a knowledge graph and checks novelty. & The knowledge graph is rebuilt for each domain from abstracts in the common corpus instead of the original biology and materials graph. \\
CycleResearcher & \href{https://github.com/zhu-minjun/Researcher/tree/8c1b253092f3f41b3efc414fb189004855bb49f8}{\nolinkurl{github.com/zhu-minjun/Researcher}} (\texttt{8c1b253}) & For each slot, ten native candidate papers, Standard-mode review by four reviewers, and Top1 selection as in the official tutorial. & Paper writing keeps the original 19,000-token output limit; an empty or truncated paper counts as a native failure. \\
AI Scientist v3 & \href{https://github.com/SakanaAI/AI-Scientist-v3/tree/c95861ee97d2bc60a61de5fe4609329482c00a7d}{\nolinkurl{github.com/SakanaAI/AI-Scientist-v3}} (\texttt{c95861e}) & Native ideation stage only (the official ideation script and instructions), without experiments or paper writing. & No retrieval or network access: each slot reads five reference papers frozen in advance from the common corpus (mostly abstracts); Semantic Scholar is not used. \\
\midrule
Single (ours) & \href{https://github.com/Einsia/InsightPro/tree/main/system}{\nolinkurl{github.com/Einsia/InsightPro}} & One agent uses the InsightPro prompt to connect problems with insights, develop mechanisms, and check prior work, then delivers Idea Cards (Section~\ref{sec:ideax}). Five isolated sessions per domain each keep only their top card. & Accesses the insights and problems in InsightPro. Web search is not used with any generator; with the generators other than Astra, InsightPro is served through a local interface. We also provide a web-enabled version, whose generated ideas are shown on our demo website. Time limit per agent turn: 4\,h by default, 24\,h for Qwen3.8-27B, and 12\,h for Qwen3.8-Flash-Next and Claude Opus. \\
Multi (ours) & \href{https://github.com/Einsia/InsightPro/tree/main/system}{\nolinkurl{github.com/Einsia/InsightPro}} & Six author agents each draft five cards (30 candidates), discuss them for one round and respond, and five cards are selected per domain. & Same as Single. \\
\end{longtable}
\endgroup

\subsection{Results with the Claude Opus Generator}
\label{app:claude-opus}
Table~\ref{tab:claude-opus} reports our two workflows with Claude Opus~5.5 as the generator, as additional results.

\begin{table}[tp]
\centering
\footnotesize
\footnotesize
\setlength{\tabcolsep}{13.66pt}
\caption{\textbf{Our workflows with the Claude Opus~5.5 generator.} Each numeric cell is \textbf{Astra / Opus} (judges); metrics and aggregation follow Table~\ref{tab:main}.}
\label{tab:claude-opus}
\vspace{2pt}
\begin{tabular}{@{}lcccccc@{}}
\toprule
 & \multicolumn{4}{c}{\textbf{Generation}} & \multicolumn{2}{c}{\textbf{IS}} \\
\cmidrule(lr){2-5}\cmidrule(lr){6-7}
Method & Contribution & Importance & Soundness & Novelty & Lit.\ novelty & Quality \\
 & (1--5) & (1--5) & (1--5) & (0--5) & (0--100) & (0--100) \\
\midrule
Single (ours) & 3.29\,/\,3.09 & 4.07\,/\,3.39 & 3.20\,/\,3.80 & 3.97\,/\,3.62 & 59.10\,/\,57.10 & 81.70\,/\,76.30 \\
Multi (ours) & 3.34\,/\,3.11 & 4.13\,/\,3.43 & 3.28\,/\,3.87 & 3.98\,/\,3.68 & 59.00\,/\,58.60 & 81.10\,/\,77.90 \\
\bottomrule
\end{tabular}
\end{table}

\subsection{Notes on the Main Table}
\label{app:main-notes}
In Table~\ref{tab:main}, bold marks the best value within each generation model, column, and judge.

CycleResearcher keeps the 19,000-token output limit of its original paper-writing step; thinking generators can spend this limit on reasoning and leave the paper empty or truncated. Such slots count as native failures (44 for Qwen3.8-27B, 14 for Qwen3.8-Flash-Next, and 155 for GLM-5.3-Flash) and are not included in its scores. Table~\ref{tab:native-failures} lists the native failures of every method with each generation model.

\begin{table}[tp]
\centering
\footnotesize
\footnotesize
\caption{\textbf{Native failures per method and generation model.} Each cell counts, out of 190 output slots, the slots where the system terminated without delivering a final idea. All scores are averaged over the successfully generated ideas, so native failures are not included in the scores.}
\label{tab:native-failures}
\setlength{\tabcolsep}{5.00pt}
\begin{tabular}{@{}lcccc@{}}
\toprule
Method & GPT-6 Astra & Qwen3.8-27B & Qwen3.8-Flash-Next & GLM-5.3-Flash \\
\midrule
IdeaScientist & 0 & 1 & 7 & 106 \\
CoI & 2 & 0 & 1 & 1 \\
AutoResearchClaw & 0 & 2 & 0 & 0 \\
Dr.~Claw & 0 & 15 & 7 & 3 \\
DeepScientist & 0 & 0 & 7 & 6 \\
ResearchAgent & 8 & 0 & 0 & 3 \\
SciAgents & 12 & 0 & 18 & 20 \\
CycleResearcher & 0 & 44 & 14 & 155 \\
AI Scientist v3 & 0 & 1 & 8 & 15 \\
\midrule
Single (ours) & 0 & 0 & 0 & 0 \\
Multi (ours) & 0 & 0 & 0 & 0 \\
\midrule
Raw-paper RAG (ablation) & 0 & 0 & 0 & 1 \\
\bottomrule
\end{tabular}
\end{table}

\subsection{Literature Scale and Input Budget}
\label{app:context-budget}

We measured PDF text lengths for a frozen sample of 180 papers from five InsightPro domain units.
Within each unit, we sampled uniformly without replacement from distinct papers with an arXiv identifier: 60 papers in knowledge distillation and 30 in each of the other four units.
We downloaded the first arXiv PDF version, extracted all pages with \texttt{pdftotext}, normalized whitespace, and counted tokens using \texttt{o200k\_base}.
These measurements include extractable text from references and appendices; images are excluded, and formula or table extraction can be lossy.

The pooled sample mean is 16,335.6 tokens, the median is 14,804.5, and the interquartile range is 11,647.3--19,393.3.
Of the sampled papers, 156/180 (86.7\%) exceed 10,000 tokens.
Using \texttt{cl100k\_base} instead gives a mean of 16,472 tokens; retaining the original extracted whitespace gives 17,182 tokens with \texttt{o200k\_base}.
Because the sample covers five domains and first arXiv versions, it indicates order of magnitude rather than the mean length of AI papers.

Using \texttt{o200k\_base}, we separately counted the complete domain exports (the serialized insights and problems) of the 38 evaluated domains.
These measurements include JSON keys, entry content, and supporting-paper identifiers.
The median export length is 51,316.5 tokens and the range is 4,762--246,483.
Table~\ref{tab:context-budget} reports the five sampled units alongside these export measurements.

\begin{table}[tp]
\centering
\footnotesize
\caption{Measured PDF sample lengths and domain exports. The final column extrapolates each domain's sample mean to its library paper count; it is an estimate, not a measured sum of all source texts.}
\label{tab:context-budget}
\setlength{\tabcolsep}{12.20pt}
\begin{tabular}{lrrrrr}
\toprule
Domain & Papers & PDF $n$ & Mean PDF tokens & Export tokens & Estimated full text \\
\midrule
Latent reasoning & 90 & 30 & 17,494 & 10,378 & 1.57M \\
Agent memory & 375 & 30 & 20,894 & 38,737 & 7.84M \\
Speculative decoding & 261 & 30 & 15,254 & 21,646 & 3.98M \\
Knowledge distillation & 1,063 & 60 & 14,975 & 116,331 & 15.92M \\
Continual learning & 1,837 & 30 & 14,422 & 145,512 & 26.49M \\
\bottomrule
\end{tabular}
\end{table}

Paper counts refer to distinct identifiers within the corresponding library construction unit and its 2016--2026 source scope.
The construction inputs include abstracts and structured, results-masked proposal annotations rather than all original PDFs, so Table~\ref{tab:context-budget} compares sizes, not information content.
The 60 knowledge-distillation PDFs contain 898,519 tokens in total, while the unit's 366 exported entries occupy 116,331 tokens.
At the pooled sample mean, a one-million-token context holds about 61 full papers.
Figure~\ref{fig:context-axis} uses these measurements: approximately 16k tokens per paper and 16M for the knowledge-distillation unit; extrapolating the same mean gives about 4.2B tokens for InsightPro's 256,227 source papers and about 160B for the schematic's 9.6M AI papers, both order-of-magnitude estimates.

\section{Additional Protocol Details}
\label{app:protocols}

\subsection{InsightPro Representation, Construction, and Accounting}
\label{app:insightpro-details}
\subsubsection{Record and Relation Definitions}
\paragraph{Family definitions.}
An insight family groups findings about the same research object and problem, preserving paper-specific mechanisms and applicability conditions in member records.
A problem family groups recurring obstacles affecting the same object, defined by its model or system, task, modality, and role.
This object-based organization brings distinct solutions to a shared problem together while separating superficially similar work on different objects.
Entry identifiers link each aggregate statement to its members and source papers.

\paragraph{Graph, node types, and properties.}
We represent these records as the attributed graph
\begin{equation}
\mathcal{G}_{L}=(V_{L},E_{L},\tau_V,\tau_E,\phi_V,\phi_E),
\qquad V_{L}=\mathcal{D}\,\cup\,\mathcal{P}\,\cup\,\mathcal{I}\,\cup\,\mathcal{Q},
\label{eq:insightpro}
\end{equation}
where $\mathcal{D}$, $\mathcal{P}$, $\mathcal{I}$, and $\mathcal{Q}$ denote domains, papers, insight families, and problem families.
Each record has exactly one node type, so these four sets do not overlap; a paper associated with several domains remains a paper node with several membership links.
The maps $\tau_V$ and $\tau_E$ specify node and edge types; $\phi_V$ and $\phi_E$ associate their properties.
Table~\ref{tab:graph-schema} gives the concrete correspondence to the released records.

\begin{table}[tp]
\centering
\small
\caption{\textbf{InsightPro node and edge schema.} Entry properties may be missing when the available source does not support them. Source edges record provenance, including conditional or conflicting evidence.}
\label{tab:graph-schema}
\setlength{\tabcolsep}{5pt}
\renewcommand{\arraystretch}{1.08}
\begin{tabularx}{\linewidth}{@{}p{.22\linewidth}X@{}}
\toprule
Node or edge type & Meaning and properties \\
\midrule
Domain $d\in\mathcal D$ & Identifier, name, research-line and direction labels; associated paper and entry counts are derived from membership. \\
Paper $p\in\mathcal P$ & Identifier, title, year, venue, source links and available abstract; structured card with claims, problem, insight, techniques, observations, assumptions, evaluation, boundaries, evidence quotations, and textual relations. Source availability and reading basis are retained where available. \\
Insight $i\in\mathcal I$ & Identifier, statement, explanatory mechanism, scope, techniques, domain/case tier, first-seen metadata, record status, and member records preserving paper-specific findings and applicability conditions. \\
Problem $q\in\mathcal Q$ & Identifier, bottleneck statement, causal explanation, scope, techniques, domain/case tier, first-seen metadata, record status, and member records preserving paper-specific obstacles and conditions. \\
\midrule
$d\xrightarrow{\mathrm{contains}}p$ & The paper belongs to the domain's source collection; a paper can belong to multiple domains. Endpoints and the membership type define the relation. \\
$d\xrightarrow{\mathrm{organizes}}u$, $u\in\mathcal I\cup\mathcal Q$ & The domain owns the insight/problem entry through its unit identifier. This relation records organization, without a learned weight. \\
$p\xrightarrow{\mathrm{source\_for}}u$ & The paper is listed among the entry's sources. Associated member records provide the local claim, mechanism, techniques, scope, success/failure conditions, and source type when available; quotations remain accessible in the paper card. \\
\bottomrule
\end{tabularx}
\end{table}

\paragraph{Relations and their interpretation.}
These three edge types define $E_L$.
For a source edge, $\phi_E$ exposes the entry's member records whose supporting-paper list contains $p$; membership edges have no additional numeric weight.
The number of distinct incoming source edges to an entry is its \texttt{paper\_count}, which measures indexed evidence breadth rather than citation impact or confidence.
Two entries can be connected through a shared paper, for example $q\leftarrow p\rightarrow i$.
This path records shared provenance; it does not establish that $i$ solves $q$.
Similarly, textual paper-card relations such as \texttt{builds\_on} and \texttt{claims\_to\_contradict} remain attributes rather than a fully resolved citation graph.
The semantic connection from a relevant insight to a proposed solution is developed during ideation.

\subsubsection{Construction and Source Coverage}

\paragraph{Domain formation and paper assignment.}
The initial catalog contains 529 research lines, organized using conference-published research areas and comparison with candidate groups derived from papers. Here, a research line is a topic characterized by a shared core problem and related findings; its preliminary finding summaries guide classification and are distinct from the final aggregated insight entries.
For the full corpus, preliminary paper descriptions of core problems and findings are embedded with Qwen3-Embedding-0.6B. A neighborhood graph combines semantic proximity with shared-reference evidence where available, and hierarchical Leiden clustering produces candidate groups for classification. An LLM compares paper titles, publication metadata, and core problems against relevant catalog entries, assigning papers by the problems they address and allowing new lines when a group shares a problem outside the candidates. Subsequent consolidation and review merge overlapping lines and split mixed topics, producing 950 research lines with recorded paper memberships.

The final construction units further consolidate small research lines. For a line with fewer than 100 papers, semantic representations retrieve up to five neighboring candidate lines. An LLM reads their core problems and representative paper-card claims, deciding whether they support a coherent account of problems and insights. A proposed merge must contain no more than 500 distinct papers; pre-existing larger lines remain available. Lines without a suitable merge retain their independence. This stage produces 856 units. Each unit's paper set is the deduplicated union of its constituent lines' memberships, matching the released domain catalog. These units are the domains reported in the main text.

\paragraph{Extraction and synthesis.}
For each paper, extraction separates its claims, problems, insights, techniques, observations, assumptions, evaluation, and boundaries, together with supporting quotations.
Unsupported fields remain empty, and abstract-only sources contribute only information stated in the abstract.
Within a domain unit, paper cards are processed chronologically against separate insight and problem buffers.
Each contribution either joins a matching family or creates a new one; member records and deduplicated supporting-paper sets preserve its provenance.

Consolidation identifies missed merges and separates families that mix different research objects.
Large collections are processed in semantically ordered batches.
Conflicts are represented through their applicability conditions, distinguishing findings from different settings from disagreements under comparable conditions.
Support-based tiering separates recurring domain-level content from less broadly supported cases, recording the breadth of evidence in the indexed corpus.

\paragraph{Access and provenance.}
The read-only interface supports domain browsing, entry lookup, lexical BM25 search, and paper access.
A complete-domain export retrieves the structured entries through pagination and records the returned count.
Source metadata identify approximately 46\% of paper cards as abstract-based, 11\% as full-text-based, and 43\% as unknown-source.
Every entry retains its source links and applicability conditions, so support can be checked, historical problems revisited, and details recovered during ideation.

\paragraph{Release formats.}
The dataset is distributed in SQLite and Parquet formats, including the domain catalog, insight/problem entries, paper metadata and structured cards, and domain--paper links.
The release includes a launcher for serving the downloaded database through the read-only retrieval API.

\subsubsection{Corpus Accounting}

Structured paper cards are available for 256,129 papers; 98 distinct papers have no card.
The source metadata contain 38 AI and related conference labels, including NeurIPS, ICML, ICLR, AAAI, CVPR, ACL, EMNLP, and ICRA.
The domain tier contains 13,924 insights and 14,264 problems; the case tier contains 39,377 insights and 37,594 problems.
These tiers do not overlap, giving 53,301 insights and 51,858 problems, or 105,159 entries in total.
Domain entries summarize recurring findings and obstacles; case entries preserve more specific evidence.
Across domains, the median counts are 198 papers, 45 insights, and 44.5 problems; the corresponding ranges are 15--3,568, 1--520, and 1--450.

Counting a paper once for each domain that contains it gives 256,361 paper--domain assignments; deduplicating by identifier gives 256,227 distinct papers, all matched to source metadata.
The papers span 2016--2026.
Figure~\ref{fig:library-statistics} reports the per-domain counts.
The 38 evaluated domains are a subset of these domains.

An aggregate family's supporting-paper list does not mean that every member supports every clause of its statement.

\subsection{Generation Scoring}
\label{app:generation-scoring}

For Generation's Contribution, Importance, and Soundness dimensions, reference material is retrieved by the proposal's problem and mechanism from a common offline corpus of raw papers and fixed before either judge scores it.
The material marks each paper as available in full text, as an abstract, or as metadata only, and contains no condensed InsightPro insights or problems; both judges receive the same material, which remains unchanged during scoring.
For conceptual Novelty, each judge scores each proposal on its own, without additional retrieval and without seeing the other proposals from the same system and domain.
Judges do not see generator identities, previous scores, or the other judge's assessments.

\subsection{Historical Reconstruction Details}

Each system submits at most ten candidate cards per task.
Single uses two five-card runs; Multi selects ten from the 30 candidates written by its authors after discussion and revision.
Raw-paper RAG generates ten in one run, whereas IdeaScientist and CoI use repeated single-proposal runs.
Historical main generators depend on date, using Opus 4.6 or 4.8; auxiliary models and corpora are not identical across all systems.
The target papers are drawn from the broad knowledge resource underlying InsightPro, and some baseline configurations have restricted components, including a missing citation graph for CoI.

\subsection{Reproducibility Artifacts}

The code repository contains the prompts and the generation and evaluation code, and the released dataset contains the InsightPro library.

\subsection{Differences from the Original IdeaScientist Metrics}
\label{app:is-metrics-adaptation}

We evaluate existing research proposals and their Blueprints without reproducing IdeaScientist's historical cutoff tasks, so we do not measure Cutoff. We report the six quality dimensions, their equal-weight mean $Q$, and literature novelty $N$. Here, $N$ corresponds only to All in the original IdeaScientist protocol: using the related literature retrieved at our evaluation time as reference, it judges what is novel about the proposal at the mechanism level. We do not compute In-domain or the composite Overall score.

Unlike the original 1--10 scoring, we map the scores of the six quality dimensions and literature novelty to a common 0--100 scale before reporting them, with $Q$ defined as the arithmetic mean of the six quality dimensions on that scale. Scores are therefore not directly comparable to those reported by IdeaScientist.

\end{document}